\documentclass{article}

    \PassOptionsToPackage{numbers, compress}{natbib}
    \usepackage[preprint]{neurips_2023}

\usepackage[utf8]{inputenc} % allow utf-8 input
\usepackage[T1]{fontenc}    % use 8-bit T1 fonts
\usepackage{hyperref}       % hyperlinks
\usepackage{url}            % simple URL typesetting
\usepackage{booktabs}       % professional-quality tables
\usepackage{amsfonts}       % blackboard math symbols
\usepackage{nicefrac}       % compact symbols for 1/2, etc.
\usepackage{microtype}      % microtypography
\usepackage[table,xcdraw]{xcolor}         % colors

\usepackage{graphicx}
\usepackage{iitem}
\usepackage{wrapfig}
\usepackage[accsupp]{axessibility}
\usepackage{amsmath}
\usepackage{amssymb}
\usepackage{arydshln}
\usepackage{bm}
\usepackage{multirow}
\usepackage{subcaption}
\usepackage{cancel}

\title{Efficient Text-Guided 3D-Aware Portrait Generation \\ with Score Distillation Sampling on Distribution}

\author{%
Yiji Cheng$^{1, \ast}$ \quad
Fei Yin$^{1, \ast}$ \quad
Xiaoke Huang$^{1}$ \quad
Xintong Yu$^2$  \quad
Jiaxiang Liu$^2$ \\
\quad
\textbf{Shikun Feng}$^2$ 
\quad
\textbf{Yujiu Yang}$^1$ 
\quad
\textbf{Yansong Tang}$^{1, \dag}$
\\
$^1$Tsinghua University \quad  $^2$Baidu Inc.
}

\newcommand{\ours}{\textbf{DreamPortrait}}

\begin{document}
\renewcommand{\thefootnote}{\fnsymbol{footnote}}
\footnotetext[1]{Equal contribution.}
\footnotetext[2]{Corresponding author.}
\renewcommand*{\thefootnote}{\arabic{footnote}}

\maketitle
%%%%%%%%% TEASER

%%%%%%%%% ABSTRACT
\begin{abstract}

Text-to-3D is an emerging task that allows users to create 3D content with infinite possibilities.
Existing works tackle the problem by optimizing a 3D representation with guidance from pre-trained diffusion models.
An apparent drawback is that they need to optimize from scratch for each prompt, which is computationally expensive and often yields poor visual fidelity.
In this paper, we propose \ours{}, which aims to generate text-guided 3D-aware portraits in a single-forward pass for efficiency.
To achieve this, we extend Score Distillation Sampling from datapoint to distribution formulation, which injects semantic prior into a 3D distribution.
However, the direct extension will lead to the mode collapse problem since the objective only pursues semantic alignment.
Hence, we propose to optimize a distribution with hierarchical condition adapters and GAN loss regularization.
For better 3D modeling, we further design a 3D-aware gated cross-attention mechanism to explicitly let the model perceive the correspondence between the text and the 3D-aware space.
These elaborated designs enable our model to generate portraits with robust multi-view semantic consistency, eliminating the need for optimization-based methods.
Extensive experiments demonstrate our model's highly competitive performance and significant speed boost against existing methods.

\end{abstract}

%%%%%%%%% BODY TEXT
\section{Introduction}

With the emergence of the generative models and the multi-modal models,
a series of works concentrate on text-guided portrait editing~\cite{patashnik2021styleclip} and generation~\cite{xia2021tedigan}.
People can obtain a vivid or imaginary image from a single prompt, which opens up infinite possibilities for creating diverse and compelling visual content. 
In this work, our concentration is on the task of text-to-3D-face by leveraging unstructured 2D images and their corresponding descriptions.

\begin{figure}[t]
  \centering
  \includegraphics[width=\textwidth]{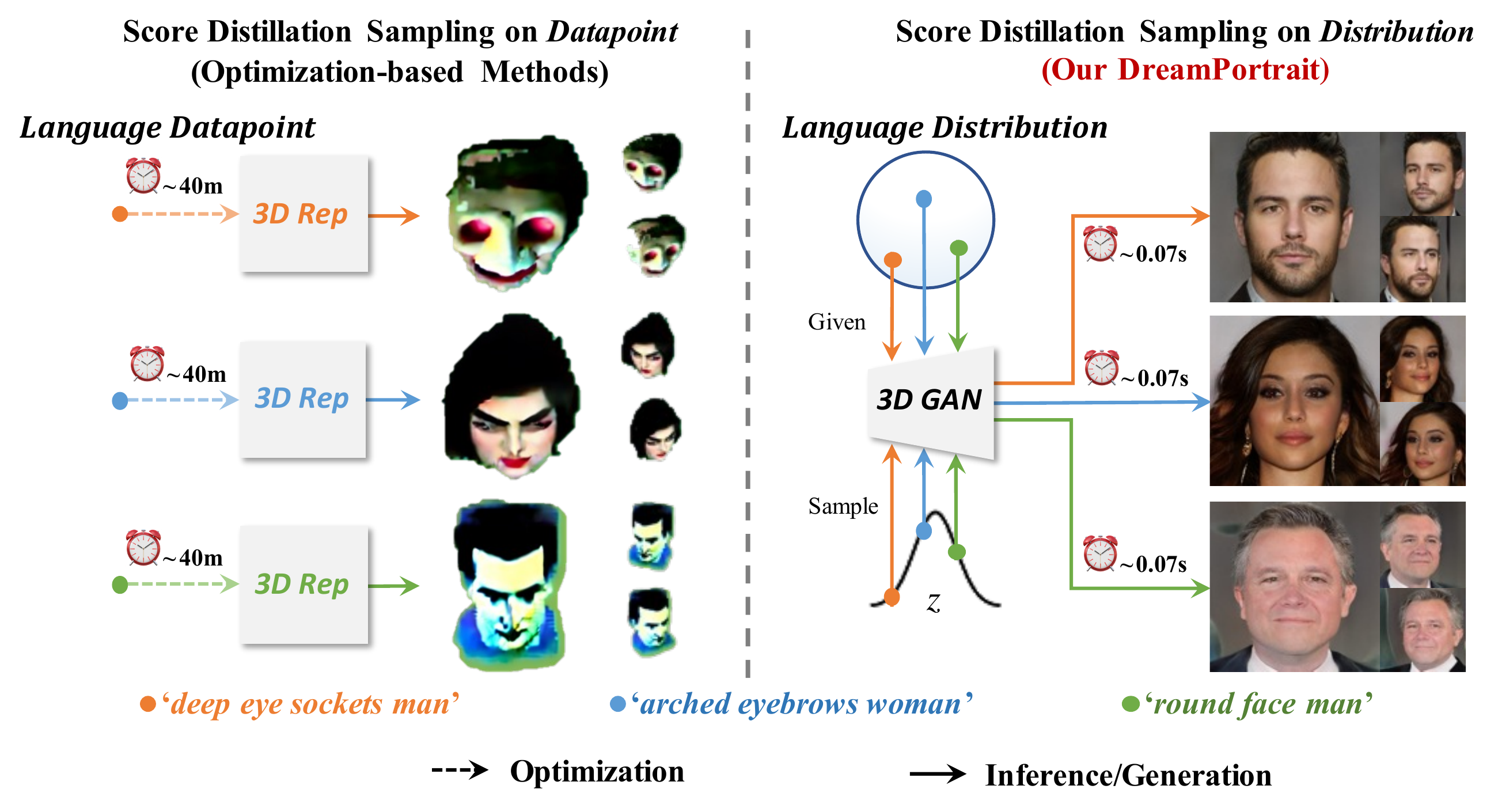}
  \caption{\small{\textit{Left}: Existing text-to-3d methods treat every prompt as a single datapoint and require optimization from scratch each time, which is computationally expensive.
  \textit{Right}:
  \ours{} (Ours) aims to map any given prompt to a 3D-aware high-fidelity portrait in a single-forward pass. 
  }}
  \label{fig:teaser}
  \vspace{-15pt}
\end{figure}

Several recent works have attempted to tackle the text-to-3D-face problem with powerful diffusion models~\cite{rombach2022ldm},
which are pre-trained on massive paired text-image data and store numerous semantic information.
Given a prompt, a line of works~\cite{poole2022dreamfusion, wang2022score, seo2023let, song2022diffusion} use pre-trained diffusion models as a discriminator to optimize an implicit 3D representation. 
Such discriminator aims to ensure that the synthesized image from 3D representation matches the semantic information of language across different views.
Another direction~\cite{kim2022datid3d, xu2022dream3d} treats the diffusion model as a pseudo generator.
The generated pseudos would provide explicit constraints for a 3D representation. 
However, a clear drawback of these methods is that they treat every prompt as a single datapoint and require optimization from scratch each time.
As shown in the left part of Fig.~\ref{fig:teaser}, they need different 3D representations to synthesize objects matching different texts, which is computationally expensive and impractical.

Another challenge in the text-to-3D-face is how to inject language information considering the 3D relationship.
Performing 3D reconstruction on the results from a 2D generator~\cite{hong2022headnerf, roich2022pivotal} may lose semantics information and lead to multi-view semantic inconsistency.
Popular 3D GANs~\cite{chan2022efficient} employ convolution layers and treat the intermediate features as plain 2D inputs, which neglects the 3D relationship, while using dense MLP to query thousands of points commonly in NeRF~\cite{mildenhall2020nerf} will significantly increase memory consumption and inference time.
Hence, a structure that efficiently counts 3D information with semantics is urgently needed.

In this paper, we propose \ours{}, which aims to achieve text-to-3D-face distribution alignment instead of datapoint optimization for efficiency.
Specifically, we propose to extend the single datapoint Score Distillation Sampling (SDS) objective~\cite{poole2022dreamfusion} to distribution form.
The distribution is possessed by a pre-trained 3D generator, which can be viewed as a 3D-aware prior.
As shown in the right part of Fig.~\ref{fig:teaser}, the optimized distribution can map text to 3D facial domain in one model, which amortizes the optimization time required for new prompts.
However, directly transferring the optimization process from point to distribution presents a clear challenge: \textit{the distribution may degenerate and raise the mode collapse problem since SDS only pursues semantics alignment}.

To prevent mode collapse, we propose two key strategies: optimizing the distribution using hierarchical condition adapters and incorporating GAN loss regularization. By tuning the adapters alongside the generator, we can gradually infiltrate the diffusion prior into the unaligned distribution in a benign manner, preventing mode collapse. Additionally, the regularization technique preserves the 3D-aware prior of the generator during the semantic alignment process.
We further design a \textit{3D-Aware Gated Cross-Attention} adapter to explicitly let the model perceive the 3D correspondence.
The adapter handles features of projections on different planes from the same volume synchronously, which enables the model to construct a more robust 3D-aware space, \textit{i.e.} the tri-plane representation space.
The obtained framework endows the 3D generator with the ability to generate portraits under any given prompt, eliminating the need for iterative forward or optimization. 
\section{Related Works}

\noindent \textbf{Text-conditioned Image Synthesis}.
The task focuses on synthesizing images that are aligned with the given textual descriptions.
There are three types of paradigms to accomplish the tasks.
GANs~\cite{reed2016generative} can unconditionally generate photo-realistic images under adversarial training, where generators are supervised by discriminators~\cite{karras2019style, karras2020analyzing, karras2021alias}.
Whereas much effort was put into improving the synthesis quality of text-conditioned GANs compared with the unconditional ones~\cite{reed2016generative, li2022stylet2i, tao2020df, reed2016learning, xu2018attngan, zhang2017stackgan, xia2021tedigan, sun2022anyface, zhou2022towards, tao2023galip, kang2023scaling, sauer2023stylegan}.
Recent Diffusion Models (DMs) have achieved amazing image synthesis quality both unconditionally and conditionally.
DMs gradually learn to de-noise the Gaussian noises to restore the real data distributions~\cite{ho2020denoising, song2020denoising}. 
The diffusion process is stable to learn and induct less mode collapse compared with GANs.
Equipped with powerful text encoders that capture rich semantics, the DMs can synthesize high-quality and versatile images that perfectly align with the input text~\cite{nichol2021glide, feng2022ernie, saharia2022photorealistic, rombach2022ldm, li2023gligen, zhang2023adding}.
The last one follows the fashion of Auto-Regression (AR), which predicts the next tokens based on the previous tokens~\cite{yu2022scaling, zhang2021ernie}.
Directly implementing 3D text-to-image generators following the 2D counterparts are infeasible.
Our work extends text-to-image generation to the regime of 3D by distilling 2D diffusion prior.

\noindent \textbf{3D-aware Synthesis}.
One might come up with the procedure for 3D-aware synthesis by directly lifting the 2D generators to 3D, \textit{i.e.}, by optimizing the 3D assets under the supervision of the multi-view 2D synthesis images.
However, the 2D generators cannot guarantee view consistency since their training does not incorporate multi-view consistency regularization.
Thus the procedure generates quality-unsatisfactory 3D assets.
Besides, it requires lengthy optimization which prohibits its usability.
We need models akin to those in 2D generation which is efficient at inference.
However, unlike 2D images, 3D representations are versatile.
There are mesh, point cloud, occupancy, Signed Distance Function (SDF), or hybrid ones among them.
Each of them can be stored either explicitly as values fields or implicitly via neural networks.
Researchers are actively working on 3D generators that adapt to certain representations.
GANs learn the 3D distributions via adversarial training.
There are two types of instantiations of GANs.
One is 3D-aware GANs which are supervised by only 2D signals~\cite{schwarz2020graf, niemeyer2021giraffe, chan2021pi, chan2022efficient, gaoGET3DGenerativeModel}.
They need to model the camera distributions in parallel to leverage volume rendering.
One renowned work is EG3D~\cite{chan2022efficient}, which proposes tri-plane neural representation for efficient 3D distribution learning and inference.
Our work leverages tri-plane representations and enables EG3D text-aligned 3D generation.
The other is 3D GANs which directly learns 3D representations by 3D supervision, \textit{e.g.} meshes~\cite{gaoTetGANConvolutionalNeural2022}, point clouds~\cite{li2018point}, and SDF~\cite{or2022stylesdf}.
This line of work mainly focuses on shape or texture alone.
Our work distills powerful 2D diffusion priors to 3D GAN generator via language guidance, which enables efficient and text-controllable 3D generation.

\noindent \textbf{Conditional 3D-aware Synthesis}.
To control the 3D generation process, researchers incorporate various controlling factors (e.g., text~\cite{nichol2022point, wu2023high, xu2022dream3d, jain2022zero}, semantic masks~\cite{park2019gaugan}, sketch~\cite{sun2023make}, layout~\cite{bahmani2023cc3d}) into 3D generator during training to align model outputs with controlling factors.
GANs leverage vision priors like segmentation, sketch, and layout as conditions~\cite{deng20233d}.
Such an approach cannot transfer to text due to the large distribution gap between text and vision representations.
Recently, DreamFace~\cite{zhang2023dreamface}, ClipFace~\cite{aneja2022clipface} and Describe3D~\cite{wu2023high} all generate text-guided 2D texture maps to render 3D morphable models.
Meanwhile, 3D diffusion models, such as Rodin~\cite{wang2022rodin}, achieve high fidelity at the cost of vast computation.
DreamFusion~\cite{poole2022dreamfusion} introduce Score Distillation Sampling (SDS) loss, an alternative method by distilling text-controlled 2D diffusion priors~\cite{rombach2022ldm} into 3D fields~\cite{poole2022dreamfusion, wang2022score, lin2022magic3d, raj2023dreambooth3d, seo2023let}.
However, it suffers from long optimization duration and Janus problem~\cite{lin2022magic3d}, and lack of fine-grained control.
Most importantly, SDS is designed for solely optimizing certain data points instead of learning distributions.
Our work combines the efficiency of 3D-aware GAN and the expressibility of 2D diffusion models.
We efficiently distill the text-guided 2D diffusion prior into the 3D distributions of pre-trained GANs, which enables fast text-controlled 3D generation in a feed-forward manner.

\section{Method}

\subsection{SDS Extension from Datapoint to Distribution}

To achieve text-to-3D synthesis, Score Distillation Sampling (SDS)~\cite{poole2022dreamfusion} has been proposed to distill textual information from a pre-trained diffusion model to a differentiable neural renderer along the gradient direction to high probability density region.
Here, the distillation teacher, \textit{i.e.}, the diffusion model $\phi$, is able to gradually transform a sample from the Gaussian noise distribution towards image distribution.
It employs a denoising function $\epsilon_\phi(x_t;y,t)$ to predict the noise $\epsilon$ given the noisy input $x_t$, noise level $t$, and corresponding text embedding $y$.
The distillation student, \textit{i.e.}, the renderer, can produce an image $x$ at the desired camera pose with $x = g(\theta)$, where $g$ is a volumetric renderer and $\theta$ is a coordinate-based MLP representing a 3D volume.
Then, the gradient objective of SDS can be formulated as follows:
\begin{equation}
    \nabla_{\theta} 
    \mathcal{L}_{\text{SDS}}
    (\phi, g(\theta)) = 
    \mathbb{E}_{t, \epsilon} \! \! \left[ w(t)(\epsilon_\phi(x_t;y,t) - \epsilon)\frac{\partial x}{\partial \theta} \right],
\end{equation}
where $w(t)$ is a weighting function. 
In practice, classifier-free guidance~\cite{ho2022classifierfree} strategy is incorporated to obtain better visual quality.

Traditional SDS attempts to obtain a 3D datapoint, \textit{i.e.}, optimizing a 3D representation to a high probability density regions conditioned on a single text embedding.
Facing new prompts, A clear drawback is that it has to optimize from scratch. 
Differently, we are interested in generating 3D avatars under any text instruction without further optimization.
An intuitive solution is to directly extend SDS loss to a distribution formulation.
We replace the 3D scene renderer with a pre-trained 3D GAN which can map a simple noise distribution $P_z$ to the 3D image domain. 
Text conditions are also extended to the distribution form $P_y$ to match the sampled latent accordingly.
The synthesized image $x$ can be obtained with $x = g(\theta, z, y)$, where $z \sim P_z$ is noise and $y \sim P_y$ is text condition.
Then, SDS can be rewritten as follows:

\begin{equation}
    \nabla_{\theta} 
    \mathcal{L}_{\text{SDS}}
    (\phi, g(\theta, z, y)) = 
    \mathbb{E}_{z, y, t, \epsilon} \! \! \left[ w(t)(\epsilon_\phi(x_t;y,t) - \epsilon)\frac{\partial x}{\partial \theta} \right]. 
\end{equation}

\begin{wrapfigure}{r}{0.4\textwidth}
    % \center
    \vspace{-10pt}
    \includegraphics[width=0.4\textwidth]{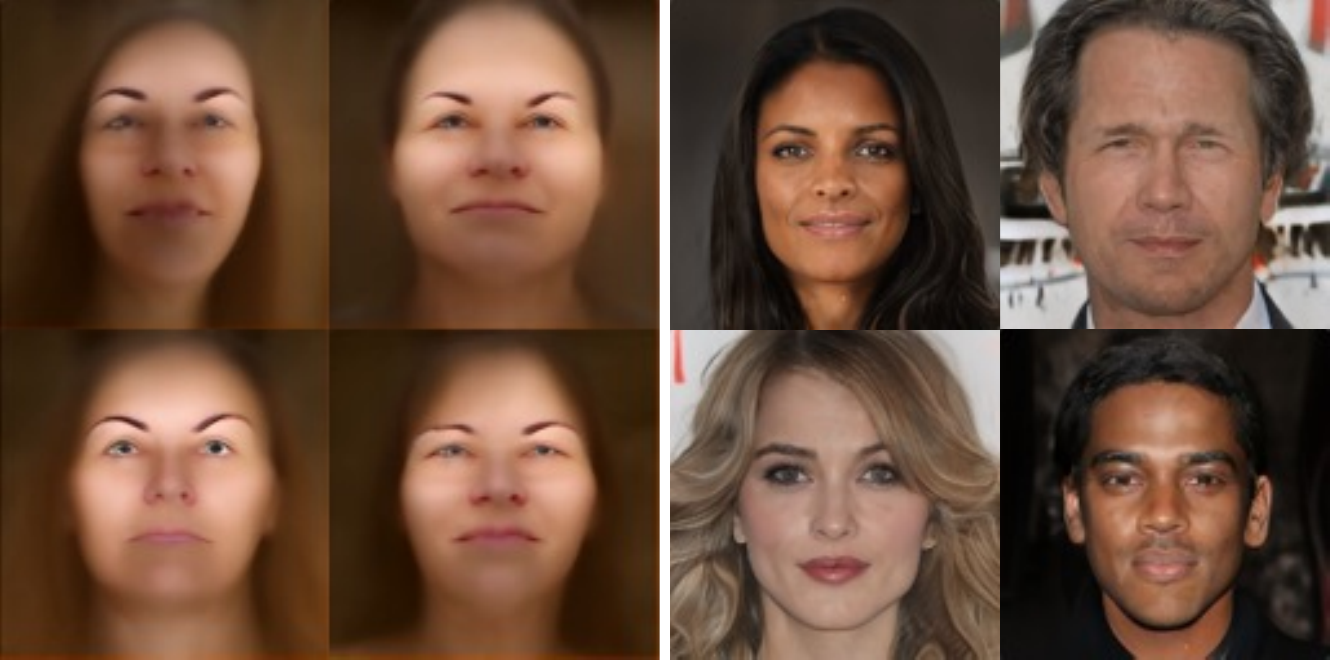}
    \caption{\small{Distribution collapsed versus maintained examples.}}
    \label{fig:collapse}
    \vspace{-25pt}
\end{wrapfigure}

Although the extended SDS objective pursues the generated image to match the sampled text prompts, it cannot guarantee the preservation of original distribution.
Direct tuning may raise mode-collapse, where the original distribution will shrink into a low-frequency point.
The generator would degenerate and synthesize coherent blur faces shown in Fig.~\ref{fig:collapse}.
\begin{figure}[t]
  \centering
  \includegraphics[width=\textwidth]{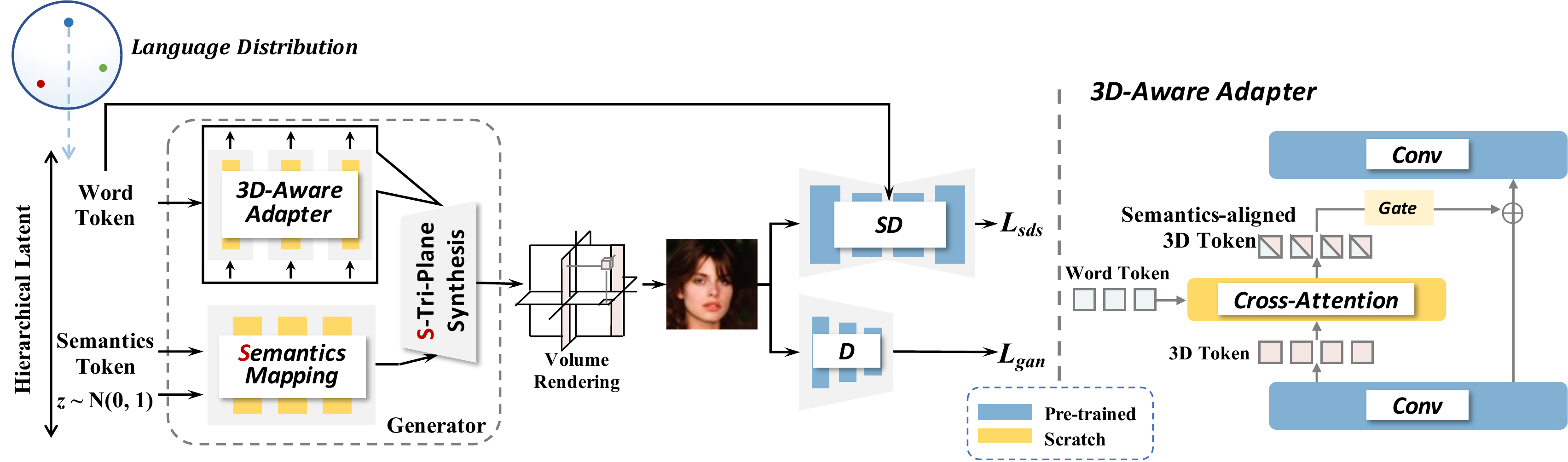}
  \caption{\small{Model overview. 
  \textit{Left}: our text-to-3D-face generator. 
  Given a prompt, we achieve different-grained tokens through language distribution and inject the information through the hierarchical latent condition. The generator synthesizes multi-view portraits. D-SDS along with GAN loss encourages robust semantics alignment.
  \textit{Right}: adapter employs 3D-aware gated cross-attention for multi-modality interaction.
  }}
\end{figure}

To address the issue of distribution collapse, we incorporate the original training objective, \textit{i.e.}, a GAN loss to discriminate the dataset images and synthesized images, as a regularization term.
The regularization can inhibit the collapse of the original 3D distribution and allow it to gradually match the text input. 
To further suppress the collapse and improve generation diversity, we propose to incorporate adapters, which is elaborated in Sec.~\ref{sec:3.2}.

\vspace{10pt}
\subsection{Hierarchical Latent Condition}
\label{sec:3.2}

The 3D generator need mapping global textual information to obtain holistic semantic consistency.
We equip the generator with a text encoder to allow for free semantic mapping during the inference stage.
Specifically, we fuse the embedding from the encoder with the generator style latent and modulate the convolution kernels.

However, simply injecting text into style latent would leave the model heading to the dense region among the language distribution and overfit the prior in the diffusion model for each prompt. 
As a consequence, the model can only present a fixed face from a prompt ignoring the influence of the style code variance.
Different prompts would also be mapped to similar identities which means the original distribution has been shrunk.
To avoid the degradation of the original 3D distribution, we incorporate adapters to inject local information.
The local adapters are inserted between the generator convolution layers, which employ the cross-attention mechanism elaborated in Sec.~\ref{sec:3.3}.
Local adapters use zero initialization~\cite{zhang2023adding, alayrac2022flamingo} strategy to achieve robust training.
Consequently, the \textit{local} adapters along with the \textit{global} mapping form the \textit{Hierarchical Latent Condition}, which significantly improves generation diversity and prevents mode collapse shown in the right part of Fig.\ref{fig:ablation-quali}.

To further let the model perceive detail changes in input prompts, we inject global and local textual information through the hierarchical latent condition respectively.
We leverage the prior from a pre-trained vision-language model, \textit{i.e.}, CLIP~\cite{radford2021learning} to extract different-grained information.
In practice, we first pad and tokenize the input prompt to achieve the corresponding vector.
Then we feed the vector to a frozen CLIP text encoder and select the penultimate layer feature~\cite{saharia2022photorealistic} as our text embedding.
The obtained embedding consists of an EOT~(end of text) token $t_\text{EOT}$ and word tokens $\{ t_\text{word}^{i} \}$ corresponding to the $i^{th}$ word in the sentence.
$t_\text{EOT}$ aggregates abundant sentence-level information, which is sufficient to guide the generation to match the overall text implication.
$\{ t_\text{word}^{i} \}$ stores detailed information which integrates a larger semantic space for variation.

Intuitively, we use $t_\text{EOT}$ for \textit{global text injection}.
Style-based generators~\cite{karras2019style, chan2022efficient} are conditioned by style latent $w$ mapped from a Gaussian noise $z$.
We concatenate $w$ and $t_\text{EOT}$ with an additional mapping network $M$ to achieve the semantics latent $w_t$.
The $w_t$ is used for modulating the generator convolution kernels to inject global guided information.
Note we did not interfere with the initial mapping process to preserve the original 3D prior.
As to \textit{local text injection}, word tokens $\{ t_\text{word}^{i} \}$ are injected through the local adapters. % for 
The cross-attention mechanism enables textual information to interact with the corresponding features at each layer.

\begin{figure}[t]
  \centering
  \includegraphics[width=\textwidth]{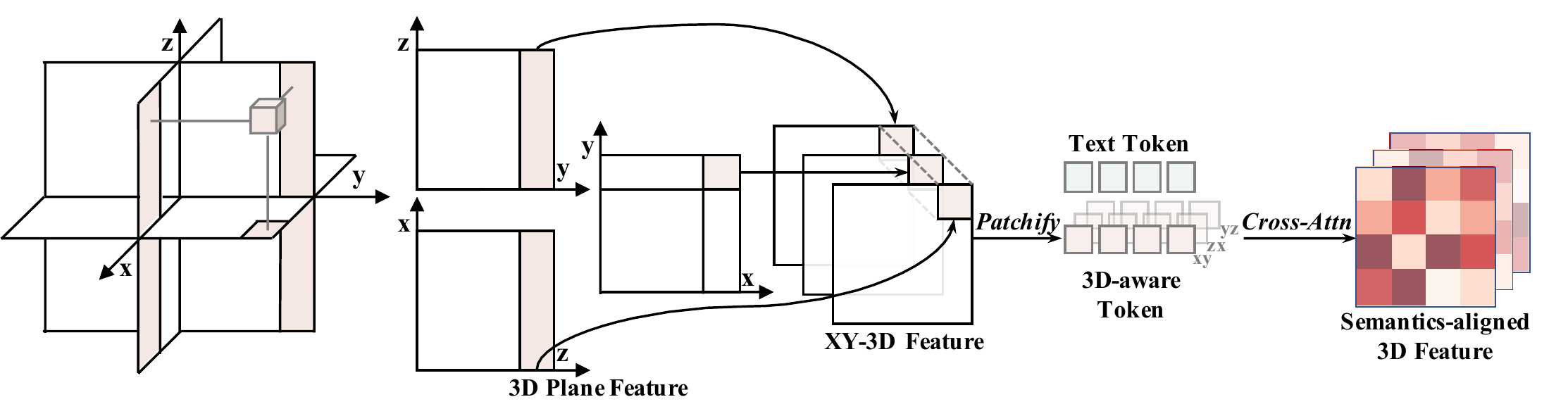}
  \caption{\small{\textit{Left}: 3D relationship of tri-planes features. \textit{Right}: 3D-aware Cross-Attention paradigm. Each tri-plane with its corresponding feature is explicitly aggregated for multi-modality attention calculation. Here we showcase $f_{xy}$ as an illustrative example.}}
  \label{fig:3d-attention}
\end{figure}

\subsection{3D-aware Gated Cross-Attention Adapter}
\label{sec:3.3}
Following recent multi-modal generative models~\cite{rombach2022ldm, kang2023scaling}, we adopt attention mechanism in adapters to align the 3D-aware space and the text domain.
Attention calculation requires converting 3D features into tokens. Simply patchifying channel-wise concatenated tri-planes will cause the mixing of theoretically uncorrected features in terms of 3D correlation~\cite{wang2022rodin}.
Hence, we propose \textit{3D-aware cross-attention} mechanism.

\noindent \textbf{3D-aware Feature}. 
When introducing text conditions using cross-attention, it is essential to establish the correct spatial relationship between text and 3D features.
As shown in Fig.~\ref{fig:3d-attention}, points in space are projected onto three orthogonal planes to query features.
Hence, a point on a feature plane accumulates a vertical column of information and corresponds to two projection lines in other planes.
In other words, the feature point along with two lines describes a 3D relationship, which should be emphasized synchronously.

Following this bias, we can obtain 3D-aware features via explicitly extracting the tri-plane with its corresponding features for multi-modality calculation. 
Specifically, tri-plane features $f$ can be decoupled as $[ f_{xy}, f_{yz}, f_{zx} ] \in \mathbb{R}^{3 \times H \times W \times C}$.
When handling the corresponding element of $f_{xy}$, we apply axis-wise pooling on other planes to aggregate two column features, \textit{i.e.}, yielding $f_{y \gets yz} \in \mathbb{R}^{H \times 1 \times C}$ and $f_{x \gets zx} \in \mathbb{R}^{1 \times W \times C}$.
Column features can be expanded to the shape of the original plane, yielding $f_{y(*) \gets yz} \in \mathbb{R}^{H \times W \times C}$ and $f_{(*)x \gets zx} \in \mathbb{R}^{H \times W \times C}$, to align with $f_{xy}$ in points.
Then, we can obtain 3D-aware features $f_{xy}^{\text{3D}}$ via concatenating these maps $[f_{(*)x}, f_{xy}, f_{y(*)}]$.
We further compute $f_{zx}^{\text{3D}}$ and $f_{yz}^{\text{3D}}$ in a similar way.
The explicit 3D-aware feature modeling greatly enhances cross-plane communication, which contributes to better alignment.

\noindent \textbf{3D-aware Gated Cross-Attention}. 
For multi-modality fusion, we first patchify the 3D-aware feature $f^{\text{3D}}$ and convert them into tokens $\{ t_{\text{3D}}^i \}$. % $f \in \mathbb{R}^{}$.
Note we set the patch size as $1 \times 1$ since we have endowed each token with corresponding 3D information. Unlike Rodin~\cite{wang2022rodin}, the restricted patch size instead of convolution can also help to evade 3D-irrelevant information.
We then perform cross-attention where $\{ t_{\text{3D}}^i \}$ is query and $\{ t_{\text{word}}^i \}$ is key and value. 

Following \cite{alayrac2022flamingo, li2023gligen}, we incorporate a gated mechanism after cross-attention to stabilize the training process.
\begin{equation}
    v = v + \beta \cdot \tanh(\gamma) \cdot \text{3D-CrossAttn}(t_{\text{3D}}, t_{\text{word}}),
\end{equation}
where $\beta$ and $\gamma$ are learnable scalars, which are initialized as $1$ and $0$ respectively.
Intuitively, the gated mechanism allows textual information to infiltrate into the model in a benign manner since the residual term is initialized as 0.
Furthermore, the gate between convolution layers can balance long-distance contexts from word tokens and local bias from convolution features.
The proposed 3D-aware gated cross-attention adapters enable the condition to access coarse-to-fine details and diverse generations.
 
\section{Experiments}
\subsection{Settings}

\noindent \textbf{Datasets}. 
We compare methods on CelebAText-HQ~\cite{sun2021multi} and FFHQ-Text~\cite{zhou2021generative}, two manually annotated real-world human face datasets.
We adopt the same data preprocessing steps as described in~\cite{chan2022efficient} and followed~\cite{sun2021multi} to split CelebAText-HQ into train and test sets. 
We further utilize a dataset, FFHQ-Text~\cite{zhou2021generative}, out of training distribution to assess the generalizability.
Each image in both datasets corresponds to nine or ten captions. 
We randomly sample a caption in the training phase and fix one caption for evaluation.

\begin{wrapfigure}{r}{0.5\textwidth}
    \center
    \vspace{-25pt}
    \includegraphics[width=0.5\textwidth]{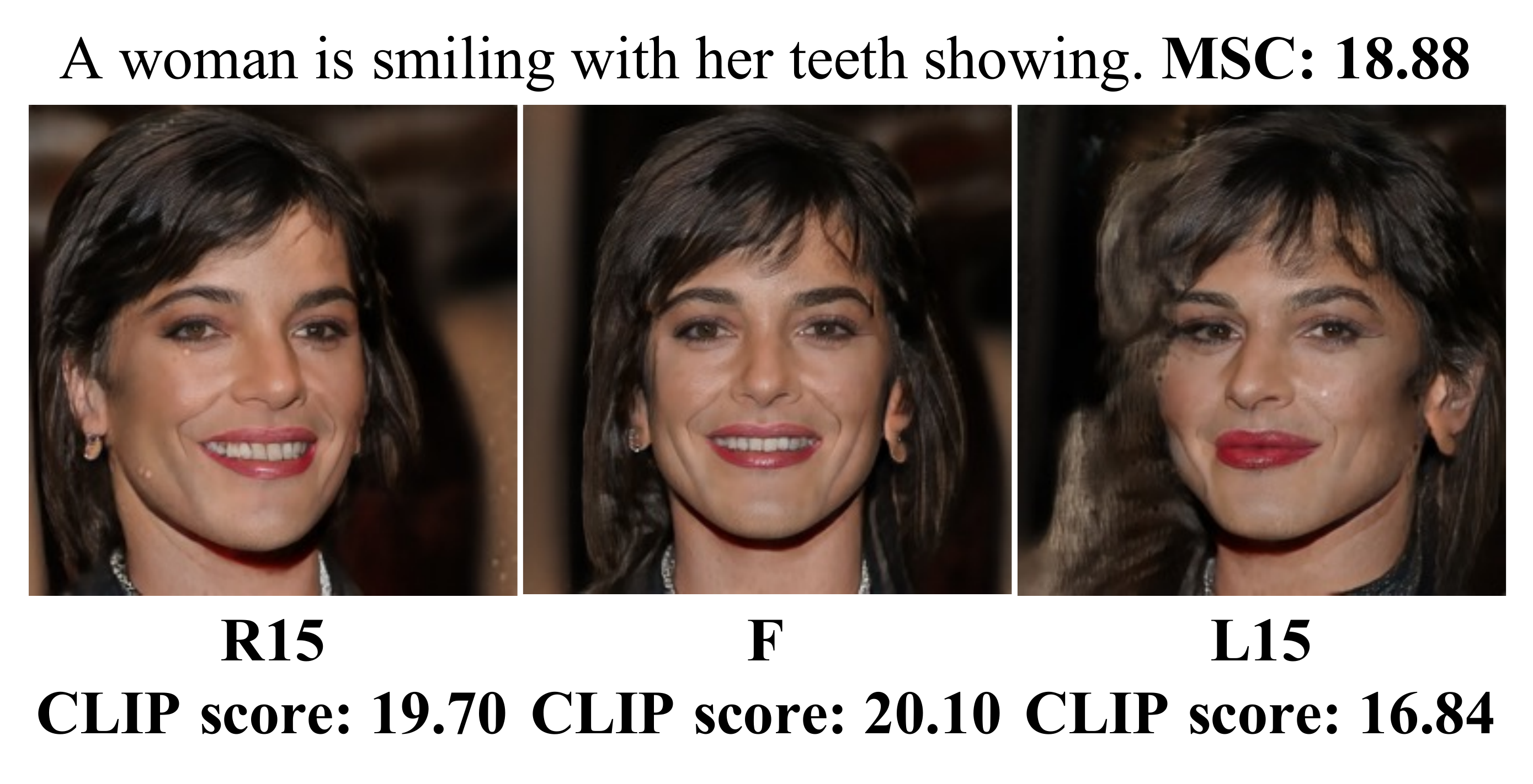}
    \vspace{-15pt}
    \caption{\small{Example of MSC. F is frontal, L is left side, R is right side, and 15 is the rotation degrees.}}
    \label{fig:msc}
    \vspace{-5pt}
\end{wrapfigure}
\noindent \textbf{Metrics}.
We compute Fr\'echet Inception Distance (FID)~\cite{heusel2017gans} and CLIP R-Precision (R-P)~\cite{park2021benchmark} to assess image quality and semantic consistency respectively.
For the task of 3D generation, it is essential to evaluate the correctness of geometry and multi-view consistency.
Hence, we further propose the Multi-view Semantic Consistency score (MSC), which is calculated as the average CLIP score of three different fixed views.
Information from different views allows the metric to better measure the 3D consistency, which is depicted in Fig.~\ref{fig:msc}.

\noindent \textbf{Baselines}.
We first compare our method against three state-of-the-art optimization-based text-to-3D methods: Stable-DreamFusion (S-DF)~\cite{poole2022dreamfusion, stable-dreamfusion}, Score Jacobian Chaining (SJC)~\cite{wang2022score}, and 3DFuse~\cite{seo2023let}.
All three methods distill semantic information from pre-trained diffusion models to a 3D representation via SDS.
To provide a comprehensive comparison, we extend our evaluation to include two-stage methods that combine text-to-2D face methods with 2D-to-3D post-processing.
We compare with TediGAN~\cite{xia2021tedigan} and StyleT2I~\cite{li2022stylet2i}, which can map text to high-resolution 2D images.
HeadNeRF~\cite{hong2022headnerf} and EG3D-based~\cite{chan2022efficient} 3D GAN inversion~\cite{yin20223dgan} method, PTI~\cite{roich2022pivotal}, are selected as dimension lifting methods.

\subsection{Comparison with Text-to-3D Methods}

\begin{figure}[t]
  \centering
  \includegraphics[width=\textwidth]{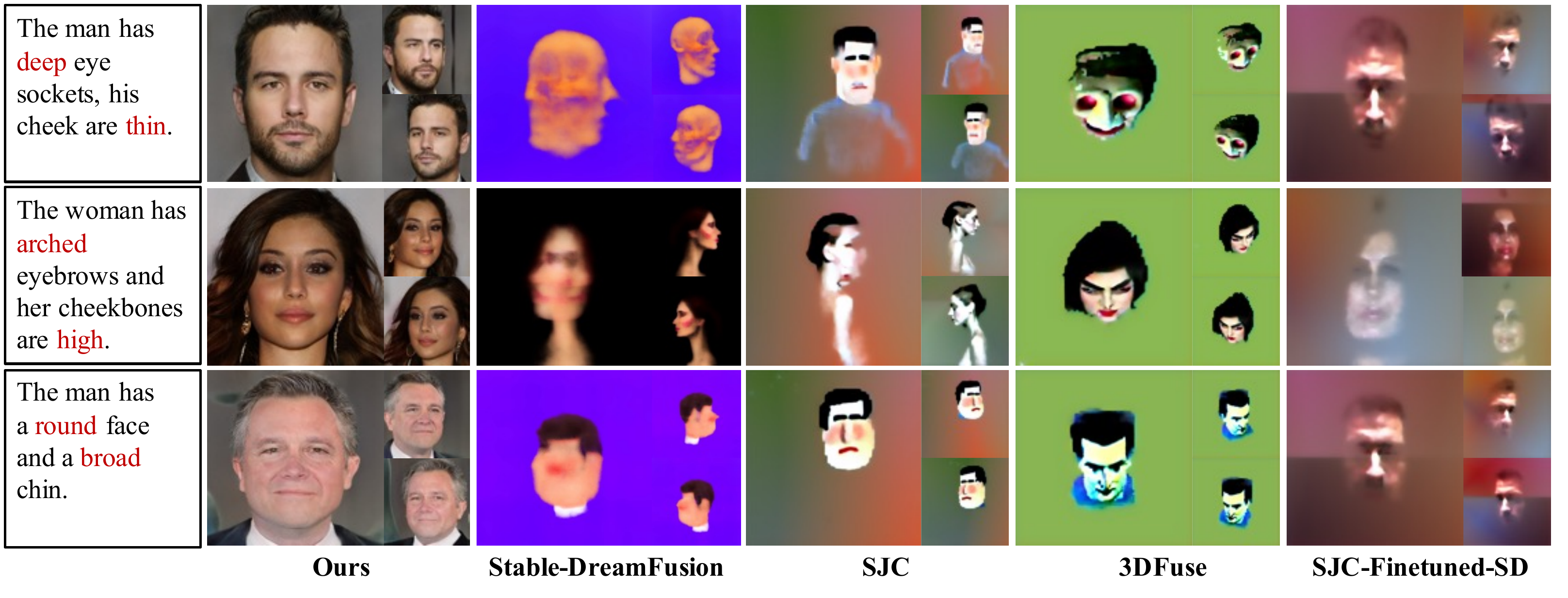} 
  \caption{\small{Qualitative comparisons with optimization-based text-to-3D methods on multi-view synthesis.}}
  \label{fig:qualitative-opt}
  \vspace{-10pt}
\end{figure}
\noindent \textbf{Qualitative Evaluations}.
In Fig.~\ref{fig:qualitative-opt}, we present a qualitative comparison with optimization-based text-to-3D methods.
Without facial prior, former methods perform poorly and generate unrealistic visual results, which can even lead to the Janus problem.
They tend to over-focus on certain parts of the words in a sentence, leading to semantic inconsistencies.
In contrast, our method gradually injects language information into a 3D distribution with the prior from pre-trained EG3D, 
resulting in high-fidelity and semantic consistent portraits. 
We also fine-tune the diffusion model on our training dataset to inject the facial prior and apply it to SJC shown in the last column.
While the face roughly appears, it suffers from severe mode collapse.

\begin{wrapfigure}{r}{0.4\textwidth}
    \center
    \vspace{-20pt}
    \includegraphics[width=0.4\textwidth]{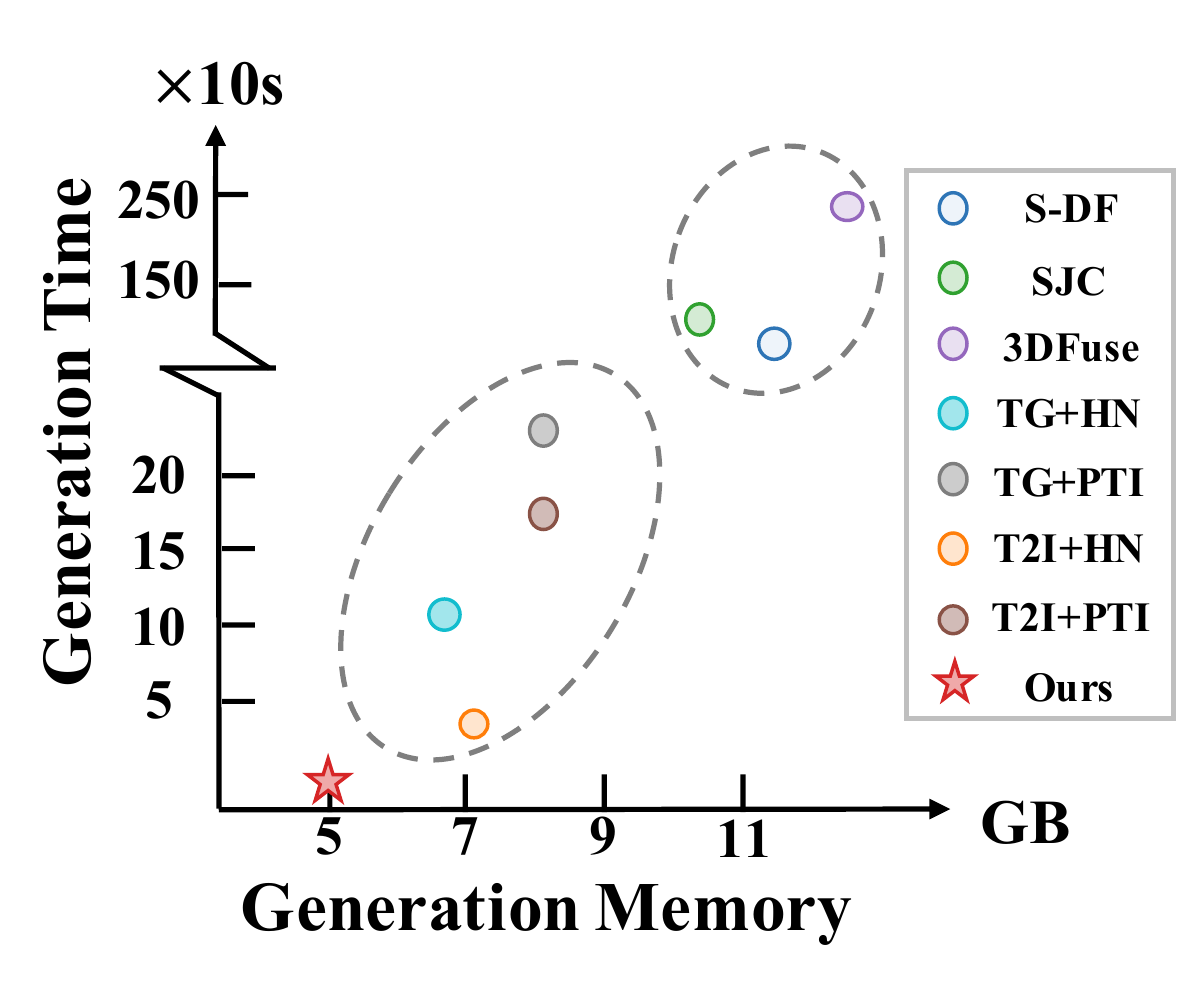}
    \vspace{-5pt}
    \caption{\small{Generation efficiency plot in terms of both memory and time.}}
    \vspace{-10pt}
    \label{fig:efficiency}
\end{wrapfigure}
\noindent \textbf{Efficiency Evaluations}.
We present the generation speed and memory efficiency of our method against baselines in Fig.~\ref{fig:efficiency}.
The metrics were measured from the moment a new text prompt was inputted until the generation of a 3D representation.
Optimization-based text-to-3D methods require training from scratch for every new text prompt, resulting in a minimum time cost of 20 minutes and a memory usage of 10 GB. 
Two-stage methods face efficiency limitations primarily in the second stage, where 3D reconstruction or 3D inversion methods are also optimization-based.
In comparison, our 3D forward method achieves a generation speed of approximately 15 FPS and only occupies about 4.5 GB of memory on a V100 GPU. These results demonstrate the efficiency of our method in generating high-quality 3D-aware portraits from text prompts at smooth, interactive frame rates.

\subsection{Comparison with Text-to-2D Lifting Methods}

\begin{table}[htbp]
\vspace{-10pt}
\centering
\caption{\textbf{Performance on CelebAText-HQ and FFHQ-Text}. 
The first four columns' results are calculated only in face regions to align with HeadNeRF.
% The results of \ours{}* are calculated with mask the region out of the face to align with HeadNeRF. 
}
\label{tab:performance-CelebAText-HQ}
\resizebox{\textwidth}{!}{%
\centering
\begin{tabular}{ll|ccc|ccc}
\toprule
\multicolumn{2}{l}{}    &\multicolumn{3}{c}{CelebAText-HQ}              &\multicolumn{3}{c}{FFHQ-Text}\\
Methods & Synthesis Type & FID$\downarrow$ & MSC$\uparrow$ & R-P$\uparrow$ & FID$\downarrow$ & MSC$\uparrow$ & R-P$\uparrow$\\ 
% \midrule
% TediGAN-base~\cite{xia2021tedigan} & 2D Inversion & 71.12 & 20.18 & 18.23 & 62.92  & 20.50 & 17.76 \\
% TediGAN-ext~\cite{xia2021towards}  & 2D Inversion & 53.15 & 20.00 & 26.92 & 55.51  & 21.05 & 19.61\\
% StyleT2I~\cite{li2022stylet2i}     & 2D Forward   & 59.45 & 18.23 & 13.00 & 138.83 & 19.64 & 10.92 \\ 
\midrule
TediGAN-base+HeadNeRF~\cite{hong2022headnerf} & 3D Reconstruction & 113.10         & 17.78          & 8.38           & 109.65         & 16.82          & 6.18           \\
TediGAN-ext\ \;+HeadNeRF                      & 3D Reconstruction & 98.66          & 18.79          & 15.07          & 106.63         & 18.25          & 10.13          \\
StyleT2I\qquad \ \ +HeadNeRF                  & 3D Reconstruction & 93.37          & 17.84          & 9.84           & 95.67          & 19.97          & 4.47           \\ 
\rowcolor{gray!10} \ours{}* (ours)                                & 3D Forward        & \textbf{17.51} & \textbf{20.66} & \textbf{35.08} & \textbf{52.82} & \textbf{21.09} & \textbf{18.02} \\ 
\midrule
TediGAN-base+PTI~\cite{roich2022pivotal} & 3D Inversion & 140.01         & 19.16          & 14.76          & 115.8          & 19.71          & 13.15          \\
TediGAN-ext\ \;+PTI                      & 3D Inversion & 112.59         & 20.21          & 22.15          & 97.12          & 21.33          & 16.97          \\
StyleT2I\qquad \ \ +PTI                  & 3D Inversion & 114.87         & 18.28          & 11.84          & 140.04         & 19.69          & 12.50          \\
\rowcolor{gray!10} \ours{} (ours)                            & 3D Forward   & \textbf{28.47} & \textbf{20.72} & \textbf{66.31} & \textbf{89.31} & \textbf{21.77} & \textbf{51.18} \\
\bottomrule
\end{tabular}%
}
\end{table}

\begin{figure}[t]
  \centering
  \includegraphics[width=\textwidth]{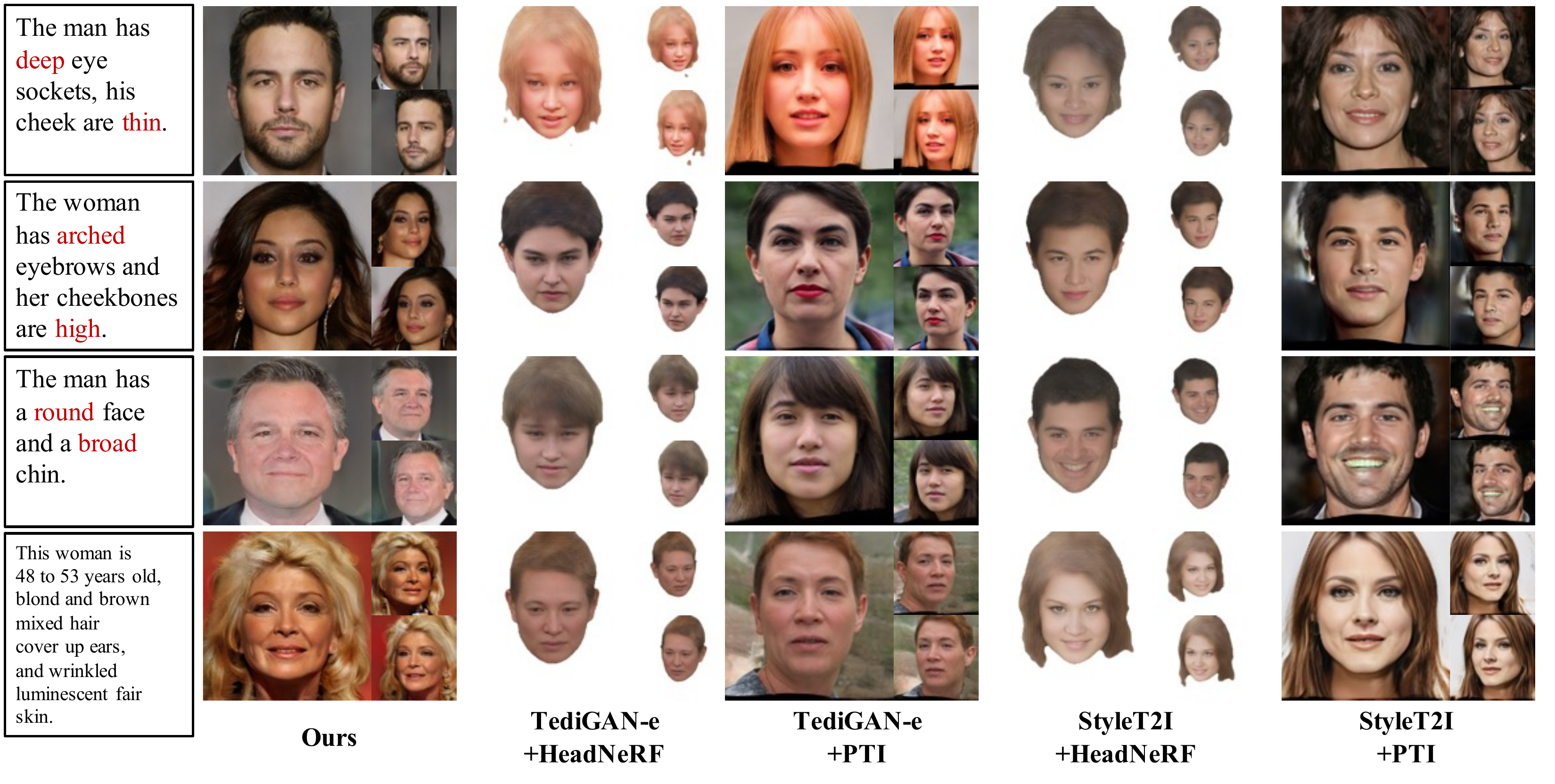}
  \caption{\small{Qualitative comparisons with two-stage text-guided generation methods on multi-view synthesis.}}
  \label{fig:qualitative-two-stage}
  \vspace{-10pt}
\end{figure}

\noindent \textbf{Quantitative Evaluations}.
Tab.~\ref{tab:performance-CelebAText-HQ} displays the quantitative comparisons between our method and baseline models. 
Our method outperforms all of the baseline models on two datasets, with a large margin on the R-P metric, which demonstrates that our method can generate high-quality 3D-aware images while maintaining multi-view semantic consistency.

Since the lifting process is an ill-posed problem and only attempts to constrain the geometry consistency, 
it cannot inpaint semantic information of unseen views. 
As a consequence, the quality of 3D faces obtained by lifting methods is lower than that of the 2D images before lifting.
Our method adopts an end-to-end learning paradigm, where language prior is injected through all modules in a 3D manner.
Ours achieves higher multi-view semantic consistency scores in both MSC and R-P.

We evaluate the generalizability on FFHQ-Text, where the text annotation significantly differs from the training samples, \textit{i.e.}, the training set of CelebAText-HQ.
Our method achieves the best semantic consistency, indicating the robust generalization ability brought by the diffusion prior.

\noindent \textbf{Qualitative Evaluations}.
Fig.~\ref{fig:qualitative-two-stage} shows the qualitative comparisons. 
Since the 3D reconstruction method HeadNeRF~\cite{hong2022headnerf} is limited to only reconstructing the head part, it undermines the background information and performs poorly on visual quality.
3D inversion methods can preserve more visual information from 2D images, but their semantic consistency is still not stable,
including the unmatching gender and accessory.
There also exists inconsistent semantic information across views, like varied makeup.
Differently, our method presents high 3D-aware fidelity along with multi-view semantic consistency corresponding to the input prompt. 

\subsection{Ablation Study}
\begin{table}[ht]
% \vspace{-10pt}
	\renewcommand{\tabcolsep}{10pt}
	% \small
	\caption{Ablations of \ours{} on CelebAText-HQ. 
 % We vary the semantics priors and 3D-aware attention to study their effects.
	}
	\begin{subtable}[!t]{0.45\textwidth}
		\centering
		\begin{tabular}{lccc}
			\toprule
			\textit{Loss} & FID$\downarrow$ & MSC$\uparrow$ & R-P$\uparrow$ \\
			\midrule
                w/o D-SDS  & \textbf{21.77} & 16.69 & 27.31 \\
			w/ CLIP  & 25.39 & 18.93 & 35.66 \\
                w/ D-SDS & 28.47 & \textbf{20.72} & \textbf{66.31} \\
			\bottomrule
		\end{tabular}
		\caption{The effect of semantics priors.}
		\label{tab:ablation-sds}
	\end{subtable}
	\hspace{\fill}
 	\begin{subtable}[!t]{0.45\textwidth}
            \centering
		\begin{tabular}{lccc}
			\toprule
			\textit{Attn.} & FID$\downarrow$ & MSC$\uparrow$ & R-P$\uparrow$ \\
			\midrule
			w/o 3D    & 28.97 & 20.59 & 65.88 \\
		    w/o gate  & \textbf{28.13} & 20.39 & 63.62 \\
                w/ both   & 28.47 & \textbf{20.72} & \textbf{66.31} \\
			\bottomrule
		\end{tabular}
		\caption{The effect of 3D-aware gated attention.}
		\label{tab:ablation-attention}
	\end{subtable}
	\label{ablations}
	\vspace{-10pt}
\end{table}
\begin{figure}[t]
  \centering
  \includegraphics[width=\textwidth]{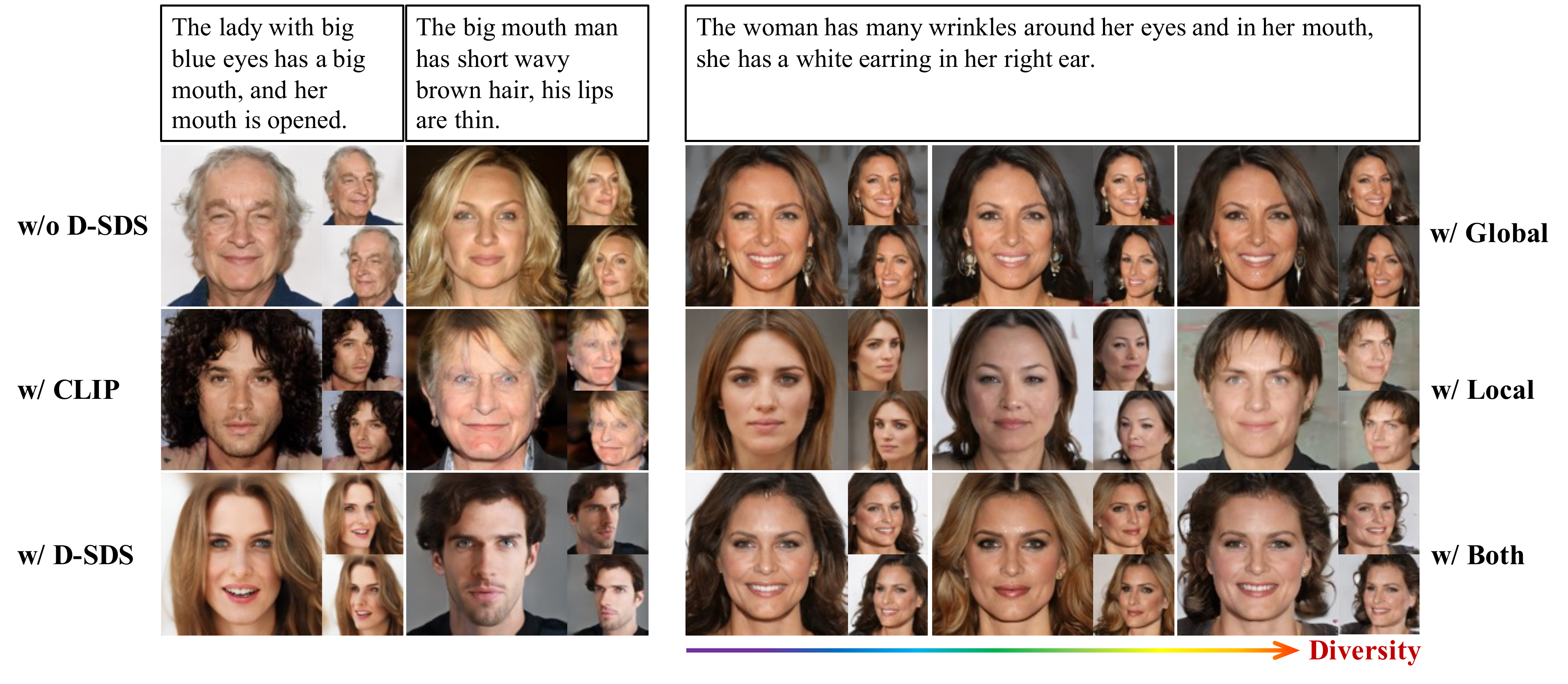}
  \caption{\small{Ablation study of semantics priors and hierarchical latent condition.}}
  \label{fig:ablation-quali}
\end{figure}

\noindent \textbf{Score Distillation Sampling on Distribution}.
We use Score Distillation Sampling on Distribution (\textbf{D-SDS}) as a means to encourage semantics alignment. 
To investigate the impact of D-SDS, we explore two options: 
(1) \textbf{w/o D-SDS}: 
similar to \cite{reed2016generative}, the language information purely stems from text embeddings;
(2) \textbf{w/ CLIP}: we replace D-SDS loss with CLIP loss to constrain generated images and corresponding prompts.
The results is shown in Tab.~\ref{tab:ablation-sds} and left Fig.~\ref{fig:ablation-quali}.
`w/o D-SDS' can hardly generate semantic consistent results and performs poorly on R-P.
`w/ CLIP' can achieve image-level semantic alignment, but it lacks the ability to provide a multi-view semantic alignment, yielding unsatisfactory multi-view semantic consistency. 
On the contrary, `w/ D-SDS' can synthesize highly semantic consistent 3D portraits, which can contribute to the pixel-level semantic alignment stored in a pre-trained diffusion model.
It is worth noting that semantic alignment would shift the original generator space, improving R-P but raising FID~\cite{sauer2023stylegan, kang2023scaling}. 
Our method achieves a trade-off without decreasing visual quality.

\noindent \textbf{Hierarchical Latent Condition}.
Recall that we utilize the hierarchical latent condition mechanism to incorporate different-grained language information, which is verified shown in right Fig.~\ref{fig:ablation-quali}.
`w/ global' injects the language information only through the global text mapper.
We can observe that it synthesizes a fixed identity given a prompt.
This is because the model overfits the prior in the diffusion model for each prompt due to the limited capacity. 
`w/ Local' can alleviate the mode collapse and generate diverse identities.
However, relying solely on `w/ Local' is  insufficient to give out a semantic-aligned 3D representation. 
`w/ Local' may over-focus on local details at the expense of global semantic matching, yielding to ambiguous results.
Therefore, a combination of both global mapper and local adapters is necessary to generate various semantics-aligned portraits.

\noindent \textbf{3D-Aware Gated Cross-Attention}.
Tab.~\ref{tab:ablation-attention} presents the results of our 3D-aware attention architectures. 
Calculating attention in a 3D-aware manner can help the model avoid introducing irrelevant information, which further increases 3D-aware semantic consistency on MSC.
The gate mechanism stables the training process and improves the image quality.

\subsection{Prompt Mixing}
\begin{wrapfigure}{r}{0.5\textwidth}
    \center
    \vspace{-25pt}
    \includegraphics[width=0.5\textwidth]{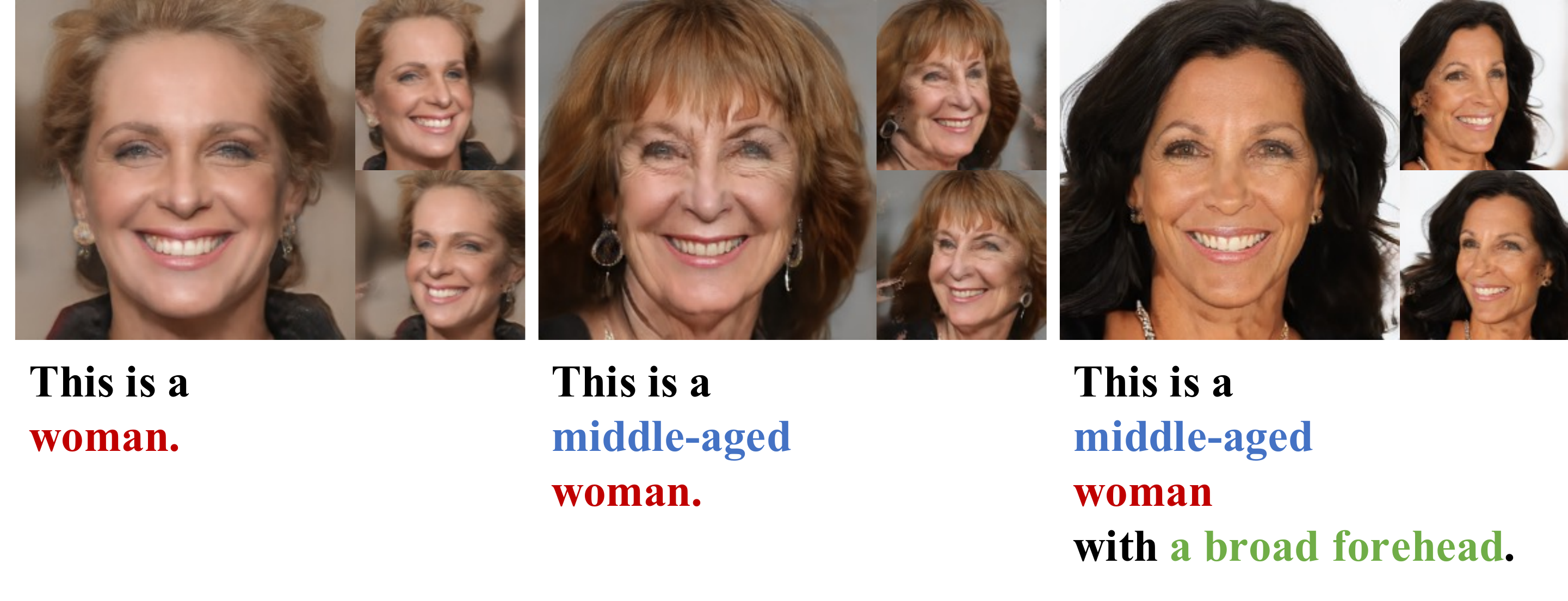}
    \vspace{-15pt}
    \caption{\small{Prompt mixing examples.}}
    \label{fig:case}
    \vspace{-50pt}
\end{wrapfigure}

Our model can perceive the slight local modification in prompts allowing for prompt mixing application.
As shown in Fig.~\ref{fig:case}, we gradually add up descriptions and the model can synthesize portraits with corresponding details, like `middle-aged' increasing wrinkles.

\section{Conclusion}
In this paper, we propose \ours{} for efficient text-guided 3D-aware portrait generation, which extends Score Distillation Sampling into distribution formulation.
By employing the hierarchical latent condition mechanism and GAN loss regularization, our method can significantly alleviate the mode collapse problem and improve generation fidelity.
Moreover, we propose the 3D-aware gated cross-attention mechanism to improve the multi-view semantic consistency.
Extensive experiments demonstrate our highly competitive performance and significant speed boost against existing methods.

\textbf{Limitation}. Our method only focuses on avatars and is not capable of handling general 3D scenes or objects.
For future work, it would be valuable to develop models with enhanced modeling capabilities and the ability to freely align with semantics, allowing for broader applications beyond avatars.

\textbf{Social Impact}. 
Text-guided portrait generation could be misused for fake images or manipulating authentic posters for improper or illegal purposes. 
These potentially harmful applications may pose a societal threat. 
We strictly forbid these kinds of abuses.

%%%%%%%%% ACKNOWLEDGEMENT
% \begin{ack}
% Use unnumbered first level headings for the acknowledgments. All acknowledgments
% More information about this disclosure can be found at: \url{https://neurips.cc/Conferences/2023/PaperInformation/FundingDisclosure}.

% Do {\bf not} include this section in the anonymized submission, only in the final paper. You can use the \texttt{ack} environment provided in the style file to autmoatically hide this section in the anonymized submission.
% \end{ack}

%%%%%%%%% REFERENCES

{\small
\bibliographystyle{ieee_fullname}
\bibliography{egbib}

\begin{thebibliography}{10}
\providecommand{\url}[1]{\texttt{#1}}
\providecommand{\urlprefix}{URL }
\providecommand{\doi}[1]{https://doi.org/#1}

\bibitem{alayrac2022flamingo}
Alayrac, J.B., Donahue, J., Luc, P., Miech, A., Barr, I., Hasson, Y., Lenc, K.,
  Mensch, A., Millican, K., Reynolds, M., et~al.: Flamingo: a visual language
  model for few-shot learning. NeurIPS  \textbf{35},  23716--23736 (2022)

\bibitem{aneja2022clipface}
Aneja, S., Thies, J., Dai, A., Nie{\ss}ner, M.: Clipface: Text-guided editing
  of textured 3d morphable models. arXiv preprint arXiv:2212.01406  (2022)

\bibitem{bahmani2023cc3d}
Bahmani, S., Park, J.J., Paschalidou, D., Yan, X., Wetzstein, G., Guibas, L.,
  Tagliasacchi, A.: Cc3d: Layout-conditioned generation of compositional 3d
  scenes. arXiv preprint arXiv:2303.12074  (2023)

\bibitem{chan2022efficient}
Chan, E.R., Lin, C.Z., Chan, M.A., Nagano, K., Pan, B., De~Mello, S., Gallo,
  O., Guibas, L.J., Tremblay, J., Khamis, S., et~al.: Efficient geometry-aware
  3d generative adversarial networks. In: CVPR. pp. 16123--16133 (2022)

\bibitem{chan2021pi}
Chan, E.R., Monteiro, M., Kellnhofer, P., Wu, J., Wetzstein, G.: pi-gan:
  Periodic implicit generative adversarial networks for 3d-aware image
  synthesis. In: CVPR. pp. 5799--5809 (2021)

\bibitem{deng20233d}
Deng, K., Yang, G., Ramanan, D., Zhu, J.Y.: 3d-aware conditional image
  synthesis. arXiv preprint arXiv:2302.08509  (2023)

\bibitem{feng2022ernie}
Feng, Z., Zhang, Z., Yu, X., Fang, Y., Li, L., Chen, X., Lu, Y., Liu, J., Yin,
  W., Feng, S., et~al.: Ernie-vilg 2.0: Improving text-to-image diffusion model
  with knowledge-enhanced mixture-of-denoising-experts. arXiv preprint
  arXiv:2210.15257  (2022)

\bibitem{gaoGET3DGenerativeModel}
Gao, J., Shen, T., Wang, Z., Chen, W., Yin, K., Li, D., Litany, O., Gojcic, Z.,
  Fidler, S.: {GET3D:} {A} generative model of high quality 3d textured shapes
  learned from images. arXiv:2209.11163  (2022)

\bibitem{gaoTetGANConvolutionalNeural2022}
Gao, W., Wang, A., Metzer, G., Yeh, R.A., Hanocka, R.: Tetgan: {A}
  convolutional neural network for tetrahedral mesh generation. In: {BMVC}.
  p.~365 (2022)

\bibitem{heusel2017gans}
Heusel, M., Ramsauer, H., Unterthiner, T., Nessler, B., Hochreiter, S.: Gans
  trained by a two time-scale update rule converge to a local nash equilibrium.
  NeurIPS  \textbf{30} (2017)

\bibitem{ho2020denoising}
Ho, J., Jain, A., Abbeel, P.: Denoising diffusion probabilistic models. NeurIPS
   \textbf{33},  6840--6851 (2020)

\bibitem{ho2022classifierfree}
Ho, J., Salimans, T.: Classifier-free diffusion guidance. arXiv preprint
  arXiv:2207.12598  (2022)

\bibitem{hong2022headnerf}
Hong, Y., Peng, B., Xiao, H., Liu, L., Zhang, J.: Headnerf: A real-time
  nerf-based parametric head model. In: CVPR. pp. 20374--20384 (2022)

\bibitem{jain2022zero}
Jain, A., Mildenhall, B., Barron, J.T., Abbeel, P., Poole, B.: Zero-shot
  text-guided object generation with dream fields. In: CVPR. pp. 867--876
  (2022)

\bibitem{kang2023scaling}
Kang, M., Zhu, J.Y., Zhang, R., Park, J., Shechtman, E., Paris, S., Park, T.:
  Scaling up gans for text-to-image synthesis. arXiv preprint arXiv:2303.05511
  (2023)

\bibitem{karras2021alias}
Karras, T., Aittala, M., Laine, S., H{\"a}rk{\"o}nen, E., Hellsten, J.,
  Lehtinen, J., Aila, T.: Alias-free generative adversarial networks. NeurIPS
  \textbf{34},  852--863 (2021)

\bibitem{karras2019style}
Karras, T., Laine, S., Aila, T.: A style-based generator architecture for
  generative adversarial networks. In: CVPR. pp. 4401--4410 (2019)

\bibitem{karras2020analyzing}
Karras, T., Laine, S., Aittala, M., Hellsten, J., Lehtinen, J., Aila, T.:
  Analyzing and improving the image quality of stylegan. In: CVPR. pp.
  8110--8119 (2020)

\bibitem{kim2022datid3d}
Kim, G., Chun, S.Y.: Datid-3d: Diversity-preserved domain adaptation using
  text-to-image diffusion for 3d generative model. In: CVPR (2023)

\bibitem{li2018point}
Li, C.L., Zaheer, M., Zhang, Y., Poczos, B., Salakhutdinov, R.: Point cloud
  gan. arXiv preprint arXiv:1810.05795  (2018)

\bibitem{li2023gligen}
Li, Y., Liu, H., Wu, Q., Mu, F., Yang, J., Gao, J., Li, C., Lee, Y.J.: Gligen:
  Open-set grounded text-to-image generation. arXiv preprint arXiv:2301.07093
  (2023)

\bibitem{li2022stylet2i}
Li, Z., Min, M.R., Li, K., Xu, C.: Stylet2i: Toward compositional and
  high-fidelity text-to-image synthesis. In: CVPR. pp. 18197--18207 (2022)

\bibitem{lin2022magic3d}
Lin, C.H., Gao, J., Tang, L., Takikawa, T., Zeng, X., Huang, X., Kreis, K.,
  Fidler, S., Liu, M.Y., Lin, T.Y.: Magic3d: High-resolution text-to-3d content
  creation. arXiv preprint arXiv:2211.10440  (2022)

\bibitem{mildenhall2020nerf}
Mildenhall, B., Srinivasan, P.P., Tancik, M., Barron, J.T., Ramamoorthi, R.,
  Ng, R.: Nerf: Representing scenes as neural radiance fields for view
  synthesis. In: {ECCV}. pp. 405--421 (2020)

\bibitem{nichol2021glide}
Nichol, A., Dhariwal, P., Ramesh, A., Shyam, P., Mishkin, P., McGrew, B.,
  Sutskever, I., Chen, M.: Glide: Towards photorealistic image generation and
  editing with text-guided diffusion models. arXiv preprint arXiv:2112.10741
  (2021)

\bibitem{nichol2022point}
Nichol, A., Jun, H., Dhariwal, P., Mishkin, P., Chen, M.: Point-e: A system for
  generating 3d point clouds from complex prompts. arXiv preprint
  arXiv:2212.08751  (2022)

\bibitem{niemeyer2021giraffe}
Niemeyer, M., Geiger, A.: Giraffe: Representing scenes as compositional
  generative neural feature fields. In: CVPR. pp. 11453--11464 (2021)

\bibitem{or2022stylesdf}
Or-El, R., Luo, X., Shan, M., Shechtman, E., Park, J.J.,
  Kemelmacher-Shlizerman, I.: Stylesdf: High-resolution 3d-consistent image and
  geometry generation. In: CVPR. pp. 13503--13513 (2022)

\bibitem{park2021benchmark}
Park, D.H., Azadi, S., Liu, X., Darrell, T., Rohrbach, A.: Benchmark for
  compositional text-to-image synthesis. In: NeurIPS (2021)

\bibitem{park2019gaugan}
Park, T., Liu, M.Y., Wang, T.C., Zhu, J.Y.: Gaugan: semantic image synthesis
  with spatially adaptive normalization. In: SIGGRAPH. pp.~1--1 (2019)

\bibitem{patashnik2021styleclip}
Patashnik, O., Wu, Z., Shechtman, E., Cohen-Or, D., Lischinski, D.: Styleclip:
  Text-driven manipulation of stylegan imagery. In: ICCV. pp. 2085--2094 (2021)

\bibitem{poole2022dreamfusion}
Poole, B., Jain, A., Barron, J.T., Mildenhall, B.: Dreamfusion: Text-to-3d
  using 2d diffusion. arXiv preprint arXiv:2209.14988  (2022)

\bibitem{radford2021learning}
Radford, A., Kim, J.W., Hallacy, C., Ramesh, A., Goh, G., Agarwal, S., Sastry,
  G., Askell, A., Mishkin, P., Clark, J., et~al.: Learning transferable visual
  models from natural language supervision. In: ICML. pp. 8748--8763. PMLR
  (2021)

\bibitem{raj2023dreambooth3d}
Raj, A., Kaza, S., Poole, B., Niemeyer, M., Ruiz, N., Mildenhall, B., Zada, S.,
  Aberman, K., Rubinstein, M., Barron, J., et~al.: Dreambooth3d: Subject-driven
  text-to-3d generation. arXiv preprint arXiv:2303.13508  (2023)

\bibitem{reed2016generative}
Reed, S., Akata, Z., Yan, X., Logeswaran, L., Schiele, B., Lee, H.: Generative
  adversarial text to image synthesis. In: ICML. pp. 1060--1069. PMLR (2016)

\bibitem{reed2016learning}
Reed, S.E., Akata, Z., Mohan, S., Tenka, S., Schiele, B., Lee, H.: Learning
  what and where to draw. NeurIPS  \textbf{29} (2016)

\bibitem{roich2022pivotal}
Roich, D., Mokady, R., Bermano, A.H., Cohen-Or, D.: Pivotal tuning for
  latent-based editing of real images. TOG  \textbf{42}(1),  1--13 (2022)

\bibitem{rombach2022ldm}
Rombach, R., Blattmann, A., Lorenz, D., Esser, P., Ommer, B.: High-resolution
  image synthesis with latent diffusion models. In: CVPR. pp. 10684--10695
  (2022)

\bibitem{saharia2022photorealistic}
Saharia, C., Chan, W., Saxena, S., Li, L., Whang, J., Denton, E.L.,
  Ghasemipour, K., Gontijo~Lopes, R., Karagol~Ayan, B., Salimans, T., et~al.:
  Photorealistic text-to-image diffusion models with deep language
  understanding. NeurIPS  \textbf{35},  36479--36494 (2022)

\bibitem{sauer2023stylegan}
Sauer, A., Karras, T., Laine, S., Geiger, A., Aila, T.: Stylegan-t: Unlocking
  the power of gans for fast large-scale text-to-image synthesis. arXiv
  preprint arXiv:2301.09515  (2023)

\bibitem{schwarz2020graf}
Schwarz, K., Liao, Y., Niemeyer, M., Geiger, A.: Graf: Generative radiance
  fields for 3d-aware image synthesis. NeurIPS  \textbf{33},  20154--20166
  (2020)

\bibitem{seo2023let}
Seo, J., Jang, W., Kwak, M.S., Ko, J., Kim, H., Kim, J., Kim, J.H., Lee, J.,
  Kim, S.: Let 2d diffusion model know 3d-consistency for robust text-to-3d
  generation. arXiv preprint arXiv:2303.07937  (2023)

\bibitem{song2020denoising}
Song, J., Meng, C., Ermon, S.: Denoising diffusion implicit models. arXiv
  preprint arXiv:2010.02502  (2020)

\bibitem{song2022diffusion}
Song, K., Han, L., Liu, B., Metaxas, D., Elgammal, A.: Diffusion guided domain
  adaptation of image generators. arXiv preprint
  https://arxiv.org/abs/2212.04473  (2022)

\bibitem{sun2022anyface}
Sun, J., Deng, Q., Li, Q., Sun, M., Ren, M., Sun, Z.: Anyface: Free-style
  text-to-face synthesis and manipulation. In: CVPR. pp. 18687--18696 (2022)

\bibitem{sun2021multi}
Sun, J., Li, Q., Wang, W., Zhao, J., Sun, Z.: Multi-caption text-to-face
  synthesis: Dataset and algorithm. In: ACMMM. pp. 2290--2298 (2021)

\bibitem{sun2023make}
Sun, Y., Wu, Q., Zhou, H., Wang, K., Hu, T., Liao, C.C., He, D., Liu, J., Ding,
  E., Wang, J., et~al.: Make your brief stroke real and stereoscopic: 3d-aware
  simplified sketch to portrait generation. arXiv preprint arXiv:2302.06857
  (2023)

\bibitem{stable-dreamfusion}
Tang, J.: Stable-dreamfusion: Text-to-3d with stable-diffusion (2022),
  https://github.com/ashawkey/stable-dreamfusion

\bibitem{tao2023galip}
Tao, M., Bao, B.K., Tang, H., Xu, C.: Galip: Generative adversarial clips for
  text-to-image synthesis. arXiv preprint arXiv:2301.12959  (2023)

\bibitem{tao2020df}
Tao, M., Tang, H., Wu, S., Sebe, N., Jing, X.Y., Wu, F., Bao, B.: Df-gan: Deep
  fusion generative adversarial networks for text-to-image synthesis. arXiv
  preprint arXiv:2008.05865  (2020)

\bibitem{wang2022score}
Wang, H., Du, X., Li, J., Yeh, R.A., Shakhnarovich, G.: Score jacobian
  chaining: Lifting pretrained 2d diffusion models for 3d generation. arXiv
  preprint arXiv:2212.00774  (2022)

\bibitem{wang2022rodin}
Wang, T., Zhang, B., Zhang, T., Gu, S., Bao, J., Baltrusaitis, T., Shen, J.,
  Chen, D., Wen, F., Chen, Q., et~al.: Rodin: A generative model for sculpting
  3d digital avatars using diffusion. arXiv preprint arXiv:2212.06135  (2022)

\bibitem{wu2023high}
Wu, M., Zhu, H., Huang, L., Zhuang, Y., Lu, Y., Cao, X.: High-fidelity 3d face
  generation from natural language descriptions. arXiv preprint
  arXiv:2305.03302  (2023)

\bibitem{xia2021tedigan}
Xia, W., Yang, Y., Xue, J.H., Wu, B.: Tedigan: Text-guided diverse face image
  generation and manipulation. In: CVPR. pp. 2256--2265 (2021)

\bibitem{xu2022dream3d}
Xu, J., Wang, X., Cheng, W., Cao, Y.P., Shan, Y., Qie, X., Gao, S.: Dream3d:
  Zero-shot text-to-3d synthesis using 3d shape prior and text-to-image
  diffusion models. arXiv preprint arXiv:2212.14704  (2022)

\bibitem{xu2018attngan}
Xu, T., Zhang, P., Huang, Q., Zhang, H., Gan, Z., Huang, X., He, X.: Attngan:
  Fine-grained text to image generation with attentional generative adversarial
  networks. In: CVPR. pp. 1316--1324 (2018)

\bibitem{yin20223dgan}
Yin, F., Zhang, Y., Wang, X., Wang, T., Li, X., Gong, Y., Fan, Y., Cun, X.,
  Shan, Y., Oztireli, C., et~al.: 3d gan inversion with facial symmetry prior.
  arXiv preprint arXiv:2211.16927  (2022)

\bibitem{yu2022scaling}
Yu, J., Xu, Y., Koh, J.Y., Luong, T., Baid, G., Wang, Z., Vasudevan, V., Ku,
  A., Yang, Y., Ayan, B.K., et~al.: Scaling autoregressive models for
  content-rich text-to-image generation. arXiv preprint arXiv:2206.10789
  (2022)

\bibitem{zhang2017stackgan}
Zhang, H., Xu, T., Li, H., Zhang, S., Wang, X., Huang, X., Metaxas, D.N.:
  Stackgan: Text to photo-realistic image synthesis with stacked generative
  adversarial networks. In: ICCV. pp. 5907--5915 (2017)

\bibitem{zhang2021ernie}
Zhang, H., Yin, W., Fang, Y., Li, L., Duan, B., Wu, Z., Sun, Y., Tian, H., Wu,
  H., Wang, H.: Ernie-vilg: Unified generative pre-training for bidirectional
  vision-language generation. arXiv preprint arXiv:2112.15283  (2021)

\bibitem{zhang2023dreamface}
Zhang, L., Qiu, Q., Lin, H., Zhang, Q., Shi, C., Yang, W., Shi, Y., Yang, S.,
  Xu, L., Yu, J.: Dreamface: Progressive generation of animatable 3d faces
  under text guidance. arXiv preprint arXiv:2304.03117  (2023)

\bibitem{zhang2023adding}
Zhang, L., Agrawala, M.: Adding conditional control to text-to-image diffusion
  models. arXiv preprint arXiv:2302.05543  (2023)

\bibitem{zhou2022towards}
Zhou, Y., Zhang, R., Chen, C., Li, C., Tensmeyer, C., Yu, T., Gu, J., Xu, J.,
  Sun, T.: Towards language-free training for text-to-image generation. In:
  CVPR. pp. 17907--17917 (2022)

\bibitem{zhou2021generative}
Zhou, Y., Shimada, N.: Generative adversarial network for text-to-face
  synthesis and manipulation with pretrained bert model. In: FG. pp. 01--08
  (2021)

\end{thebibliography}
}

%%%%%%%%% APPENDIX

\clearpage
\appendix

%%%%%%%%% APPENDIX

\section*{Appendix}
% Thank you for reading our supplementary materials!

\section{Implementation Details}

\begin{figure}[h]
\centering
\begin{subfigure}{0.4\textwidth}
\centering
% \fbox{\rule[-.5cm]{0cm}{4cm} \rule[-.5cm]{4cm}{0cm}}
\includegraphics[width=\textwidth]{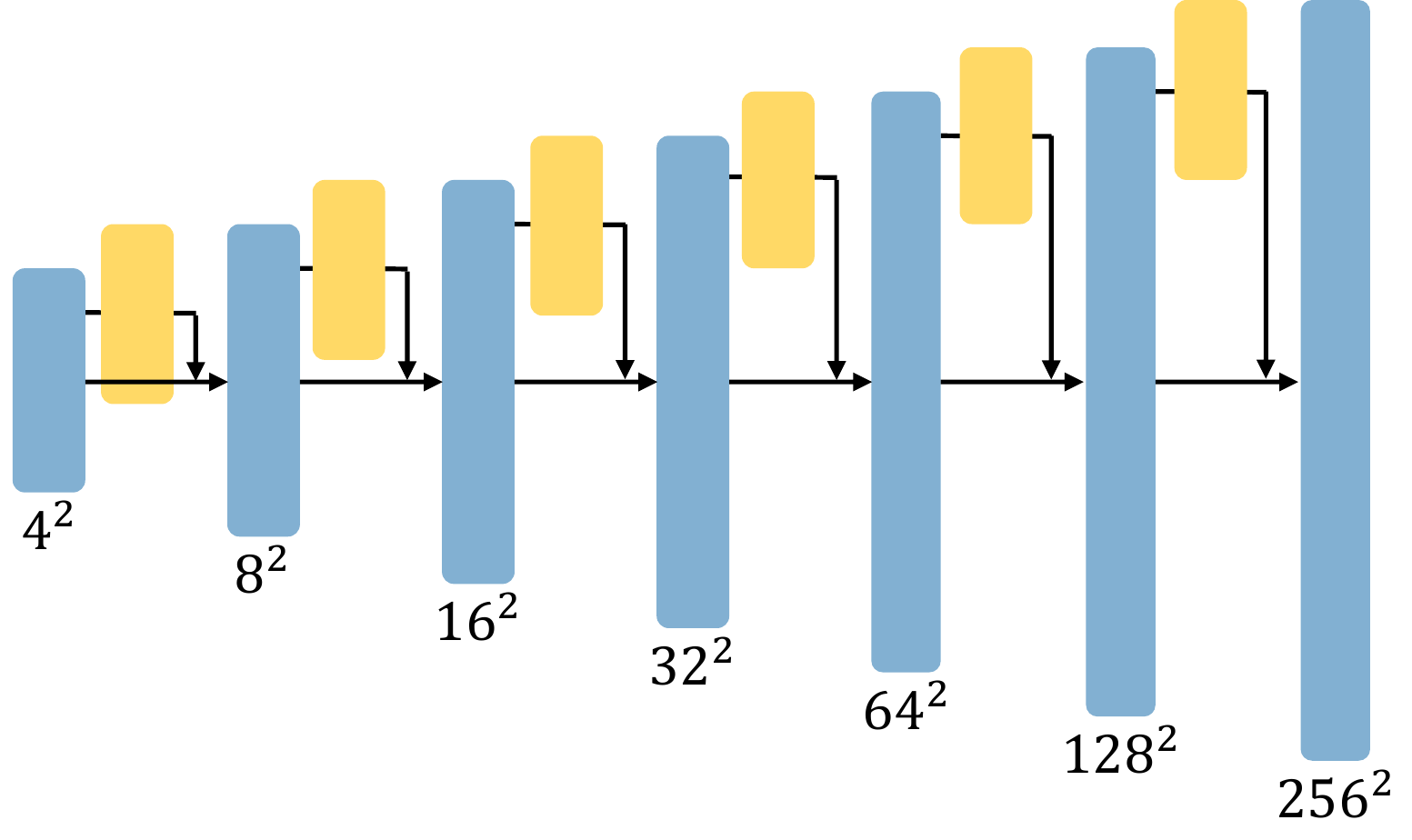}
\caption{Architecture of Synthesis Network.}
\label{fig:supp-syn}
\end{subfigure}
\begin{subfigure}{0.5\textwidth}
\centering
% \fbox{\rule[-.5cm]{0cm}{4cm} \rule[-.5cm]{4cm}{0cm}}
\includegraphics[width=\textwidth]{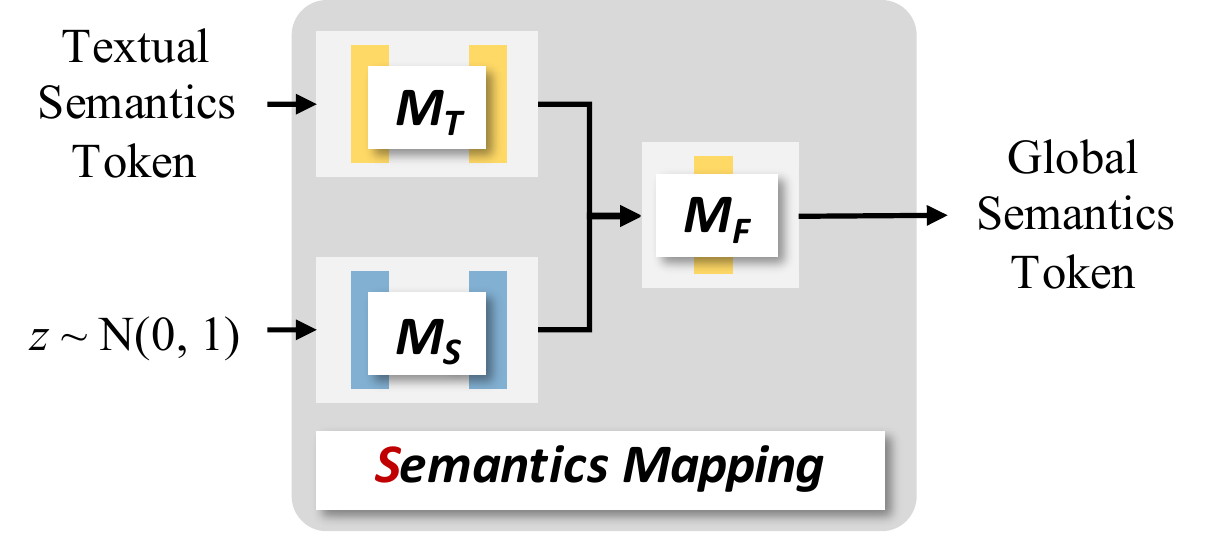} 
\caption{Architecture of Semantics Mapping Network.}
\label{fig:sup-map}
\end{subfigure}
\label{fig:supp-arch}
\end{figure}

\noindent \textbf{Backbone}.
The backbone of our approach, \textit{i.e.}, the 3D GAN, is implemented using the official PyTorch implementation of EG3D~\cite{chan2022efficient} \footnote{https://github.com/NVlabs/eg3d}.
We directly initialize the style mapping network, synthesis network, and dual discriminator with the pre-trained parameters from EG3D.
This initialization allows us to leverage the well-established 3D-aware distribution provided by EG3D as a strong starting point for our model.

\noindent \textbf{3D-Aware Adapter}.
Our 3D-aware adapter serves as a plug-and-play module, facilitating the integration of text information into the 3D-aware distribution.
Specifically, we incorporate six adapters in a residual manner between multi-scale convolution layers, covering a range from  $4^2$ to $256^2$, as depicted in Fig.~\ref{fig:supp-syn}.
This design enables an effective and seamless fusion of two modalities at different scales.

\noindent \textbf{Semantics Mapping Network}.
To ensure the unwarping of both the latent space~\cite{karras2019style} and the semantics space, our semantics mapping network is designed to accommodate two modalities inputs. 
Following EG3D~\cite{chan2022efficient}, we adopt a cascaded MLP as the structure of our semantics mapping network. 
Specifically, the network consists of two mapping layers for textual semantics tokens, two mapping layers for style codes, and an additional layer for fusing textual semantics and style, as illustrated in Fig.~\ref{fig:sup-map}.
This architecture facilitates the creation of a 3D-aware semantics space, enabling the effective combination of new prompts during generation.

\noindent \textbf{Distribution Score Distillation Sampling (D-SDS)}.
We incorporate Stable Diffusion 2.0-base~\cite{rombach2022ldm} as our diffusion prior.
Following DreamFusion~\cite{poole2022dreamfusion}, we exclude the U-Net Jacobian term as shown in Eq.~\ref{equ:unet}.
\begin{equation}
    \nabla_{\theta} 
    \mathcal{L}_{\text{SDS}}
    (\phi, g(\theta, z, y)) = 
    \mathbb{E}_{z, y, t, \epsilon} \! \! \left[ w(t)(\epsilon_\phi(x_t;y,t) - \epsilon) \bcancel{\frac{\partial x_t}{\partial x}} \frac{\partial x}{\partial \theta} \right]. 
    \label{equ:unet}
\end{equation}
Since Stable Diffusion is a latent diffusion model, we include an Auto-Encoder Jacobian term during practical implementation as shown in Eq.~\ref{equ:auto-encoder}.
\begin{equation}
    \nabla_{\theta} 
    \mathcal{L}_{\text{SDS}}
    (\phi, g(\theta, z, y)) = 
    \mathbb{E}_{z, y, t, \epsilon} \! \! \left[ w(t)(\epsilon_\phi(x_t;y,t) - \epsilon) \bcancel{\frac{\partial x_t}{\partial x'}} \frac{\partial x'}{\partial x} \frac{\partial x}{\partial \theta} \right],
    \label{equ:auto-encoder}
\end{equation}
where, $x'$ is the auto-encoding latent of $x$.

Additionally, to address the issue of sample diversity in the distribution learned through D-SDS,
we rescale the predicted noise and Gaussian noise.
This rescaling strategy is necessary to prevent potential overflow raised by the predicted noise, ensuring a stable generation.

\noindent \textbf{Data Processing}.
Following EG3D, 
we process our dataset with camera pose estimation, face alignment, and image cropping.
However, we do not mirror the images to avoid altering the semantics. 
Since each image in both datasets is associated with nine or ten captions, we randomly select one caption during the training phase and fix a specific caption for evaluation. 

\noindent \textbf{Training Strategies}.
We employ several hyperparameters and training strategies from EG3D, including training initialization with blurred real images, pose-conditioned strategy, density regularization, and learning rate adjustment.
To enhance the training stability, we introduce a two-stage strategy.
In the first stage, we only use GAN loss for training,  gradually adapting the local adapters and the global mapping network to the pre-trained EG3D.
In the second stage, we incorporate both the Distribution Score Distillation Sampling loss and GAN loss regularization to align 3D-aware semantics.
The training process involves $500K$ images in the first stage and $1000K$ images in the second stage.
We use a minibatch size of 32 and perform gradient accumulation during the second stage.
All models are trained using 8 Tesla V100 GPUs for approximately 8+48 hours.

\section{More Results}
\begin{table}[ht]
% \vspace{-10pt}
	\renewcommand{\tabcolsep}{10pt}
	% \small
	\caption{Ablations of \ours{} on CelebAText-HQ with less iterations. 
 % We vary the semantics priors and 3D-aware attention to study their effects.
	}
	\begin{subtable}[!t]{0.45\textwidth}
		\centering
		\begin{tabular}{lccc}
			\toprule
			\textit{Param.} & FID$\downarrow$ & MSC$\uparrow$ & R-P$\uparrow$ \\
			\midrule
                Freeze G      & 357  &  16.34  & 25.31 \\
			Freeze D      & 247  &  18.93  & 44.69 \\
                Unfreeze Both & 33   &  20.26  & 57.38 \\
			\bottomrule
		\end{tabular}
		\caption{The effect of trainable parameter.}
		\label{tab:supp-ablation-fix}
	\end{subtable}
	\hspace{\fill}
 	\begin{subtable}[!t]{0.45\textwidth}
            \centering
		\begin{tabular}{lccc}
			\toprule
			\textit{Num.} & FID$\downarrow$ & MSC$\uparrow$ & R-P$\uparrow$ \\
			\midrule
			2   & 58  & 20.29  & 53.94  \\
		    4   & 41  & 20.42  & 57.31 \\
                6   & 33  & 20.46  & 57.38 \\
			\bottomrule
		\end{tabular}
		\caption{The effect of 3D-aware adapter layers.}
		\label{tab:supp-ablation-layer}
	\end{subtable}
	\label{tab:supp-more-ablations}
	\vspace{-10pt}
\end{table}

\begin{wrapfigure}{r}{0.4\textwidth}
    % \center
    \vspace{-10pt}
    \includegraphics[width=0.4\textwidth]{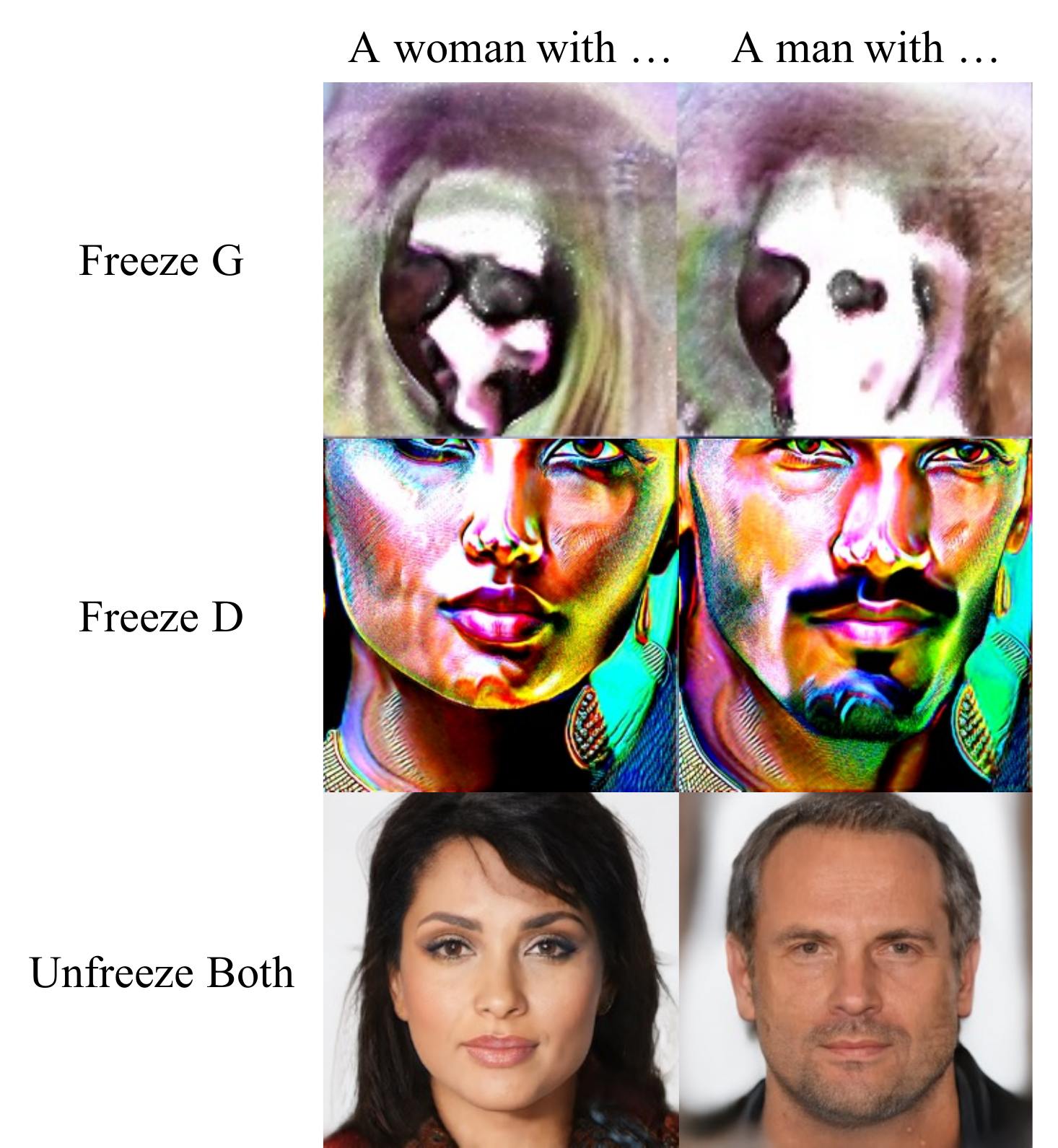}
    \caption{\small{The effect of parameter unfreezement.}}
    \label{fig:supp-fix}
    \vspace{-25pt}
\end{wrapfigure}
\noindent \textbf{Parameters Freezement of Pre-trained Model}.
We unfreeze all pre-trained parameters in our training except the diffusion model. 
From Tab.~\ref{tab:supp-ablation-fix} and Fig.~\ref{fig:supp-fix}, we observe that both freezing G and Freeze D got an exceptional high FID, which means they fail to synthesize photorealistic results.

\noindent \textbf{Number of Adapters}.
We perform an ablation study on the influence of the number of 3D-aware adapters inserted between convolution layers.
As shown in Table~\ref{tab:supp-ablation-layer}, the results demonstrate that increasing the number of adapters in the generator leads to improved metric performance.

\begin{figure}[t]
  \centering
  % \fbox{\rule[-.5cm]{0cm}{4cm} \rule[-.5cm]{4cm}{0cm}}
  \includegraphics[width=\textwidth]{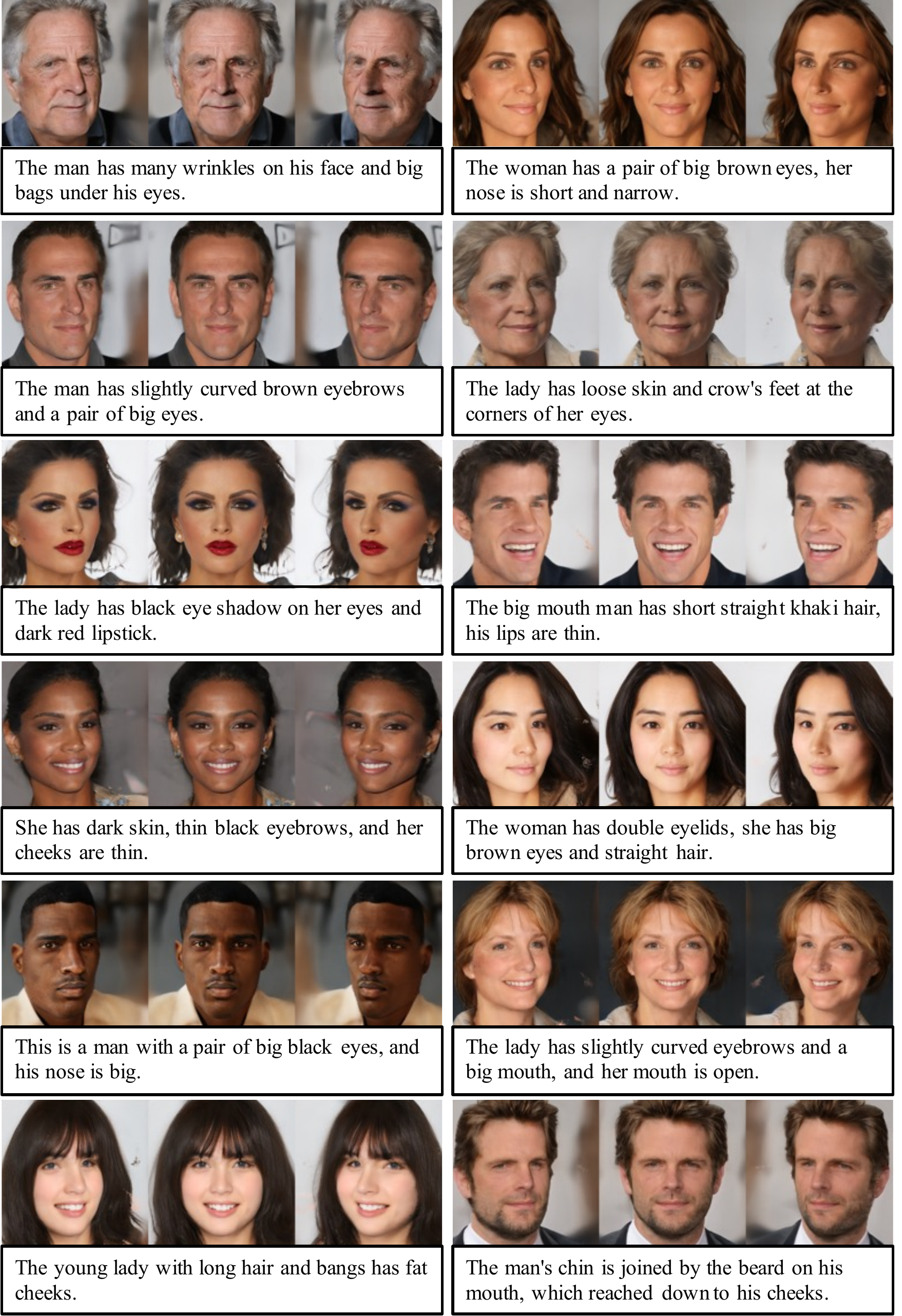}
  \caption{\small{Additional visual results.}}
  \label{fig:supp-more-comp}
  \vspace{-10pt}
\end{figure}
\noindent \textbf{Style Interpolation}.
Our generator exhibits the ability to interpolate between style latent while preserving semantics, illustrating the disentanglement latent space obtained from the semantics mapping network.
The disentangled nature allows for smooth style transitions, as shown in the attached video.

\noindent \textbf{Additional Results}.
Extra visual results can be found in Fig.~\ref{fig:supp-more-comp} and the attached video, further demonstrating the effectiveness of our model.

%%%%%%%%% REFERENCES

% \clearpage
% {\small
% \bibliographystyle{ieee_fullname}
% \bibliography{egbib}
% }

% \section*{Appendix}
% Thank you for reading our supplementary materials!
%%%%%%%%%%%%%%%%%%%%%%%%%%%%%%%%%%%%%%%%%%%%%%%%%%%%%%%%%%%%

\end{document}